\documentclass[10pt,journal,compsoc]{IEEEtran}
\ifCLASSOPTIONcompsoc
  \usepackage[nocompress]{cite}
\else
  \usepackage{cite}
\fi
\ifCLASSINFOpdf
\else
\fi
\usepackage{times}
\usepackage{graphicx}
\usepackage{amsmath}
\usepackage{amssymb}
\usepackage{adjustbox}
\usepackage{multicol}
\usepackage{multirow}
\usepackage{subcaption}
\usepackage{dsfont}
\usepackage{bbding}
\usepackage{comment}
\usepackage{xcolor}
\usepackage{algorithm}
\usepackage{algpseudocode}
\usepackage[pagebackref=true,breaklinks=true,letterpaper=true,colorlinks,bookmarks=false]{hyperref}

\newcommand{\bb}[1]{\boldsymbol{#1}}
\newlength\savewidth\newcommand\shline{\noalign{\global\savewidth\arrayrulewidth
  \global\arrayrulewidth 1pt}\hline\noalign{\global\arrayrulewidth\savewidth}}

\begin{document}
%
\title{Out-of-Distribution Detection with \\ Hilbert-Schmidt Independence Optimization}
%
%
%
%

\author{Jingyang~Lin, Yu~Wang, Qi Cai, Yingwei~Pan,~\IEEEmembership{Member, IEEE}, Ting~Yao,~\IEEEmembership{Senior Member, IEEE},\\ Hongyang Chao,~\IEEEmembership{Member, IEEE}, Tao~Mei,~\IEEEmembership{Fellow,~IEEE}
\IEEEcompsocitemizethanks{
\IEEEcompsocthanksitem Jingyang~Lin and Hongyang Chao are with SUN YAT-SEN University, China (email: yung.linjy@gmail.com; isschhy@mail.sysu.edu.cn).
\IEEEcompsocthanksitem Yu~Wang is with Qiyuan Lab, Beijing, China (email: feather1014@gmail.com).
\IEEEcompsocthanksitem Qi Cai, Yingwei~Pan, Ting~Yao, Tao~Mei are with JD AI Research, China (email: cqcaiqi@gmail.com; panyw.ustc@gmail.com; tingyao.ustc@gmail.com; tmei@live.com).}
}

%
%

\markboth{}%
{Shell \MakeLowercase{\textit{et al.}}: Bare Demo of IEEEtran.cls for Computer Society Journals}
%



\IEEEtitleabstractindextext{%
\begin{abstract}
Outlier detection tasks have been playing a critical role in AI safety. There has been a great challenge to deal with this task. Observations show that deep neural network classifiers usually tend to incorrectly classify out-of-distribution (OOD) inputs into in-distribution classes with high confidence. Existing works attempt to solve the problem by explicitly imposing uncertainty on classifiers when OOD inputs are exposed to the classifier during training. In this paper, we propose an alternative probabilistic paradigm that is both practically useful and theoretically viable for the OOD detection tasks. Particularly, we impose statistical independence between inlier and outlier data during training, in order to ensure that inlier data reveals little information about OOD data to the deep estimator during training. Specifically, we estimate the statistical dependence between inlier and outlier data through the Hilbert-Schmidt Independence Criterion (HSIC), and we penalize such metric during training. We also associate our approach with a novel statistical test during the inference time coupled with our principled motivation. Empirical results show that our method is effective and robust for OOD detection on various benchmarks. In comparison to SOTA models, our approach achieves significant improvement regarding FPR95, AUROC, and AUPR metrics. Code is available: \href{https://github.com/jylins/hood}{https://github.com/jylins/hood}.
\end{abstract}

}

\maketitle

\IEEEdisplaynontitleabstractindextext

%
\IEEEpeerreviewmaketitle

\IEEEraisesectionheading{\section{Introduction}\label{sec:intro}}
Out-of-distribution (OOD) detection, as its name implies, aims to differentiate the samples lying in a significantly different distribution from the known training data. This task is especially important for AI applications requiring high standard safety and reliability, such as medical image recognition and autonomous driving. However, However, it is not easy to tell whether or not the models ``know what they do not know''. Observations show that deep neural network classifiers usually tend to mistakenly make high confidence predictions on out-of-distribution data with wrong classification decisions into the in-distribution classes~\cite{choi2019waic,hendrycks2018deep,nalisnick2018do,Shafaei2019,pan2020exploring,zhang2016sparse,scheirer2012toward,scheirer2014probability}. The work in~\cite{RenLiklihood2019} claims that the wrong inference result during test time might be attributed to the dominant population-level background statistics shared between the in-distribution and OOD test data. Some works instead blame this to the miscalibration of the deep neural network~\cite{DuImplicit2019,Grathwohl2020Your,maaloe2019biva}. In contrast, other works found that the correctness of the statistical test method plays a more pivotal role than the network miscalibration~\cite{david2020OOD}.

In parallel to existing OOD detection frameworks, we reveal that statistical independence from in-distribution (inliers) data can be a good indicator characterizing the OOD (outliers) data. We follow the discriminative modeling assumption as in \cite{hendrycks2018deep}, where we assume access to inliers labels and OOD training data exposure during training. The goal is to differentiate test data having different classes than training inliers. We then implicitly impose the parameters' uncertainty on OOD data by penalizing the statistical dependence via Hilbert-Schmidt Independence Criterion (HSIC) between the inliers and outliers. By definition, mutual information I(X; Y) = 0 satisfies: I(X; Y) = 0 if and only if X and Y are independent. Therefore, under our new objective, the deep feature extractor on in-distribution data would refrain itself to reveal limited mutual information on OOD classes. The learned network parameters therefore become more agnostic on OOD classes as training proceeds. Note that our HOOD method can also be conveniently trained even if ground-truth real OOD training data becomes unavailable. Under this circumstance, HOOD still consistently performs better in apple-to-apple comparisons than those OOD detection methods under the same training setup, and can sometimes even perform better than those methods trained with OOD data exposure.

\begin{figure}
        \includegraphics[width=0.49\textwidth]{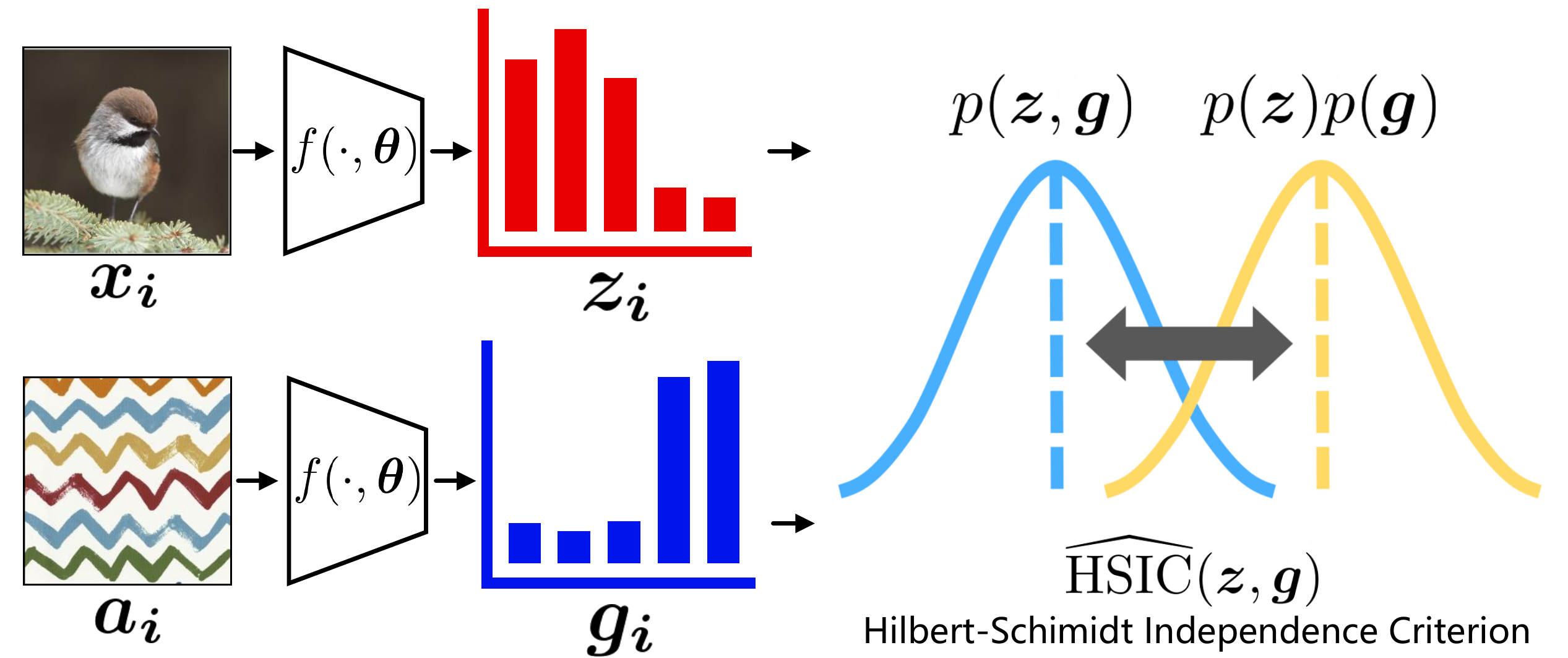}
    \caption{The motivation of the proposed HOOD framework. Our goal is to train a feature encoder $f(\cdot, \bb{\theta})$ so that the deep estimator $\bb{\theta}$ appropriately removes statistical dependence between inlier sample $\bb{x}_i$ and outlier sample $\bb{a}_i$. We adopt Hilbert-Schmidt Independence Criterion (HSIC) to measure the dependence between the inlier feature $\bb{z}$ and the outlier feature $\bb{g}$.}
    \label{fig:intro}
    \vspace{-1em}
\end{figure}

Our contribution to this paper can be summarized as follows: we reformulate the OOD detection problem from purely a statistical independence perspective. This new perspective suggests that we can model the training by resorting to a practically accessible independence measurement: HSIC metric. We also developed a simple test method during inference time, which is particularly suitable to associate with our proposed training objective. The whole training and testing procedure is named HOOD (HSIC assisted OOD detection). In Fig.~\ref{fig:intro}, we present the primary motivation for proposing HOOD. We empirically verify the advantage of applying HOOD to the OOD detection tasks on several important benchmarks, and HOOD shows strong superiority in terms of FPR95, AUROC, and AUPR metrics.

\section{Related Work}
 {\bf Generative models for OOD detection}. One mainstream paradigm for OOD detection tasks assumes no OOD data exposure during training~\cite{lee2018training,Serra2020Input,nipslikelihoodregret}. Such methods usually firstly train a generative model on the training data and then predict the generative likelihood of test data given the obtained probability density estimate. Samples lying near the low-density area are arbitrated as OOD data. However, in \cite{choi2019waic,hendrycks2018deep,nalisnick2018do,Shafaei2019}, observations show that deep generative models trained on image datasets can frequently assign a high likelihood to OOD inputs. The work in \cite{RenLiklihood2019} finds that the dominant background statistics between the in-distribution data and OOD data can drastically confound the test time inference, leading to difficulty discriminating the two groups of data. Other works discuss from the perspective of network miscalibration \cite{DuImplicit2019,Grathwohl2020Your,maaloe2019biva}. Research also investigates issues regarding data {\emph{typicality}} and the correctness of the test, which argues that samples in a high dimensional distribution usually will fall on a {typical set} with a high likelihood, which would interfere with the test \cite{nalisnick2018do,david2020OOD}.

{\bf Statistical test and uncertainty measurement for OOD detection}. Research also focuses on building alternative confidence test and statistical test to differentiate OOD data under different hypotheses \cite{AminiEvidential2020,hein2019relu,hendrycks17baseline,GeneralizedODIN,liang2020enhancing,lin2021mood,Liuenergy2020,Muratevidental2018}. The Maximum Softmax Probability (MSP)~\cite{hendrycks17baseline} method is motivated by the observation that OOD examples tend to obtain lower maximum softmax probabilities, and implement the inference based on this observation. In \cite{liang2020enhancing}, authors propose a softmax score based test assuming that in-distribution data are more sensitive to softmax temperature scaling and noise perturbations. Built upon the idea of ODIN~\cite{liang2020enhancing}, the Generalized ODIN (G-ODIN)~\cite{GeneralizedODIN} further defines a specific input pre-processing procedure and introduces a new decomposed confidence scoring test. These two additional strategies wavies ODIN from the hyperparameters tuning.  In \cite{Mdistance2018}, the authors assume that OOD data are intrinsically far away from in-distribution classes, and therefore they implement the OOD detection based on the Mahalanobis distance of each test sample to its closest in-distribution class-specific cluster center. In \cite{AminiEvidential2020,Muratevidental2018}, hierarchical probabilistic models are shown to be more effective in modeling the uncertainty of OOD data. Recent work \cite{Liuenergy2020} introduces an energy-based score for OOD detection, which shows superior performance than using the vanilla softmax score as in \cite{Mdistance2018}.  \cite{ReActSun2021} reveals an insightful observation that owing to the mismatched BatchNorm statistics, OOD data can trigger abnormal unit activations with larger variance. Inspired by this observation, \cite{ReActSun2021} proposes an activation rectification strategy, namely ReAct, which utilizes activation truncation during inference.  In this paper, we develop an alternative statistical test strategy during inference time based on the independence assumption, which is specifically suitable to associate with our proposed training objective.

{\bf Discriminative models for OOD detection}. Another line of works assumes access to in-distribution data labels \cite{fort2021exploring,hein2019relu,hendrycks2018deep,Liuenergy2020,Li2020CVPRenergy} and defines the data as having different classes than training inliers to be OOD data. This differs from what is defined in generative modeling, and the supporting hypothesis behind is that the discrimination ability between inliers and training outliers can transfer to unseen OOD test data. Examples include \cite{hendrycks2018deep}, where they explicitly impose the ``uniform label'' on OOD training data to force ``uncertainty'' on OOD classes. In \cite{Liuenergy2020}, an energy-based score is proposed to replace the softmax score in OE~\cite{hendrycks2018deep} and is shown to be empirically superior. The work in \cite{Li2020CVPRenergy} further discusses the re-sampling strategy from the OOD training data in order to improve OOD detection performance. Recent work~\cite{Yang2021semantic} further considers the impact of noisy OOD training data in practice. To mitigate the detrimental impact owing to these noisy training outliers, \cite{Yang2021semantic} introduces a refurbished benchmark based on existing OOD datasets and proposes the ``semantically coherent OOD detection'' approach. Our HOOD method also assumes access to training outliers (without their annotation information) during training. In the meanwhile, we also show that HOOD can be conveniently implemented even if we lose access to these ground-truth real training outliers by instead resorting to generated fake OOD training data.

{\bf OOD data generation.} The outlier exposure based methods heavily rely on access to training OOD data, which is impractical in real scenarios. Thus, several existing works are motivated to generate OOD training samples, which waives the need to collect real outliers requiring prior training data knowledge. The work in \cite{lee2018training} leverages Generative Adversarial Networks (GAN)~\cite{goodfellow2014generative} to generate effective training OOD samples, i.e., the ``boundary samples'' in the low-density area. Similarly, \cite{vernekar2019out} adopts conditional variational auto-encoder (CVAE)~\cite{sohn2015learning} to generate ``effective'' OOD samples, which is expected to cover the entire in-distribution ``boundary''. As mentioned in ~\cite{hein2019relu,RenLiklihood2019,sinha2021negative,tack2020csi}, strong augmentation is a cheap but effective way to generate outliers, including shifting transformation~\cite{tack2020csi}, image permuting~\cite{hein2019relu,RenLiklihood2019}, and other augmentation techniques. Our proposed HOOD also benefits from strong augmentation when real OOD training data becomes unavailable. In particular, we simply generate fake OOD training data by applying strong distortion to inliers in the pixel space when training OOD is unaccessible. Moreover, we also employ strong augmentation to expand training OOD distribution, further improving OOD detection performance.

 {\bf Comparisons to existing work.} Unlike existing literature, HOOD explores a novel hypothesis: statistical independence between in-distribution and out-of-distribution data is vital for successful OOD detection. We expect that in-distribution data should be leaking the least knowledge about OOD data, i.e., the mutual information between inliers and outliers should be small. We show that the statistical independence with inliers is a stronger property characterizing OOD samples given the current prevailing OOD detection evaluation metrics. Instead of the prevailing softmax thresholding policy, we also propose a simple statistical test metric termed COR during inference time, which is specifically suitable to couple with our training objective. We also empirically demonstrate that the proposed test metric is clearly superior to softmax thresholding under the HOOD framework. Rather than directly applying strong augmentation to generate OOD samples, we investigate the influence of strong augmentation on various OOD detection methods. In the meanwhile, we also explore the impact of different augmentation strengths. We show that HOOD encourages better learning capability and generalization ability.

\section{Method}

\subsection{Problem formulation}
We respectively use variable $\bb{x}$ to denote in-distribution (inlier) data and define variable $\bb{a}$ to refer to out-of-distribution (outlier) data. Assuming we have an in-distribution training data set $\mathcal{D}_{tr,in}$, in which each training data is a pair of $(\bb{x}_i, y^{in}_i)$, $i\in \{1,...,N\}$ indexes samples. Here $\bb{x}_i$ is a sample of variable $\bb{x}$, and the $i^{th}$ sample (e.g., a raw image) in $\mathcal{D}_{tr,in}$. Notation $y^{in}_i \in \{1,...,C\}=\mathcal{Y}_{in}$ is the label of $\bb{x}_i$. We define outlier samples to be a pair of $(\bb{a}_i, y^{out}_i)$, $i\in \{1,...,M\}$, where labels $y^{out}_i \in \{C+1,...,C+C'\}=\mathcal{Y}_{out}$. We define $\bb{z}_i=f(\bb{x}_i, \bb{\theta}) \in \mathbb{R}^d$ as the features of $\bb{x}_i$ out of the deep feature extractor $f(\cdot, \bb{\theta})$ and $\bb{g}_i=f(\bb{a}_i, \bb{\theta})  \in \mathbb{R}^d$ as outlier features of sample $\bb{a}_i$. Here, $\bb{\theta}$ denotes the parameters to be learned.

It is critical to firstly understand the OOD problem objective that HOOD serves. For standard discriminative models, such as \cite{hendrycks2018deep,Mdistance2018,liang2020enhancing,Liuenergy2020}, the support of outlier class distribution $p(y^{out}_i)$ and inlier $p(y^{in}_j)$ for all $i,j$ are defined to be disjoint, i.e., $\mathcal{Y}_{in}\cap\mathcal{Y}_{out}=\varnothing$. This is intrinsically a different definition from that of generative models such as \cite{Serra2020Input,david2020OOD,nipslikelihoodregret}, where the assumption instead is that the OOD data $\bb{a}_i$ are generated from a different distribution $p(\bb{a})$ other than $p(\bb{x})$. Our HOOD is one of discriminative OOD detection approaches that follows the former definition and the associated evaluation metrics, where HOOD seeks to discriminate any OOD data having different categories than training data.

\subsection{Motivation}
Many mainstream OOD detection frameworks, regardless whether OOD training data is available or not, usually rely on statistical test based on softmax score or its variants~\cite{liang2020enhancing,hendrycks2018deep,Liuenergy2020}. However, softmax is notorious for miscalibration. In fact, softmax score and the likes essentially squash {\emph{the relative distance}} of the test sample w.r.t. a class into a simplex, given the comparative value against other classes. Softmax is therefore sub-optimal to reflect statistical confidence~\cite{Muratevidental2018}, as it is basically a point estimate. Many other distance-based metrics, such as Mahalanobis-distance \cite{Mdistance2018} also resort to relative distance of an OOD sample against class centers.

In this paper, our model follows the standard discriminative OOD detection definition~\cite{hendrycks2018deep,Mdistance2018,liang2020enhancing,Liuenergy2020}, where the OOD samples are defined as different classes from the training data. Particularly, we propose and investigate a new hypothesis: the statistical independence in between the inlier data $\bb{x}$ and outlier data $\bb{a}$ can indicate OOD classes. Our training strategy therefore aims to explicitly encourage the features $\bb{g}_i=f(\bb{a}_i, \bb{\theta})$ of the OOD data to be independent of inlier feature $\bb{z}_i=f(\bb{x}_i, \bb{\theta})$, while simultaneously respecting the supervision of known inliers.

Considering the connection between statistical dependence and mutual information instantiates the goal of HOOD. Note that by definition, mutual information $I (X;Y)=0$ satisfies: $I (X;Y)=0$ if and only if X and Y are independent. Suppose we expect the deep parameters $\bb{\theta}$ to be capable of differentiating OOD samples. In that case, observing in-distribution data $\bb{x}$ through its features $\bb{z}_i=f(\bb{x}_i, \bb{\theta})$ should leak small knowledge about $\bb{a}$'s feature $\bb{g}_i=f(\bb{a}_i, \bb{\theta})$, i.e., the mutual information between features $\bb{z}$ and $\bb{g}$ should be more constrained. Otherwise, large dependence between $\bb{z}$ and $\bb{g}$ will interfere with the OOD differentiation, owing to the larger mutual information and semantic dependency between $\bb{z}$ and $\bb{g}$. Such interpretation of independence is hence inherently suitable for the motivation of discriminative OOD detection tasks. If we are confident that OOD classes represent different semantics of inliers, i.e., $\mathcal{Y}_{in}\cap\mathcal{Y}_{out}=\varnothing$, the learning on $\bb{x}$ should expose limited mutual information to those OOD classes $\mathcal{Y}_{out}$ represented by outliers $\bb{a}$.

The above incentive leads to our transparent but principled framework HOOD: we explicitly penalize the statistical dependence between $\bb{x}$ and $\bb{a}$ through applying the Hilbert-Schmidt Independence Criterion ({\bf{\emph{HSIC}}}) \cite{Grettonhsic2005}. We name the whole procedure as HOOD (Hilbert-Schmidt Independence Criterion assisted OOD detection).  While it is generally challenging to empirically estimate mutual information between variables, HOOD offers convenient and practical implementation to realize the alternative independence assumption.

\subsection{Statistical independence for OOD detection}
Recall that $\bb{z}_i=f(\bb{x}_i, \bb{\theta})$ is the deep feature extracted from $\bb{x}_i$. Our goal is to train the deep feature extractor $f(\bb{x}_i, \bb{\theta})$ such that the $\bb{\theta}$ mostly removes dependence between inlier $\bb{x}$ and outlier $\bb{a}$. We define any feature pairs $(\bb{z}_i, \bb{g}_i),...,(\bb{z}_j, \bb{g}_j$) to be a series of independent samples drawn from joint probability $p(\bb{z}, \bb{g})$.

It has been proved and guaranteed in \cite{Grettonhsic2005} that Hilbert-Schmidt Independence Criterion (HSIC) is {\emph{a sufficient dependence criterion under all circumstances}}, i.e., HSIC is $0$ if and only if the random variables are independent. For any pair of variable $\bb{z}$ and $\bb{g}$, HSIC is defined as a probability divergence between two probability distribution  $p(\bb{z},\bb{g})$ and the product of marginals $p(\bb{z})p(\bb{g})$ \cite{Grettonhsic2005,LiKernMax}:
{\small
\begin{align}
    \text{HSIC}(\bb{z},\bb{g}) & =\text{MMD}(p(\bb{z},\bb{g})||p(\bb{z})p(\bb{g}))                                                                            \\
                         & =\|\mathbb{E} [   \phi(\bb{z})\phi(\bb{g})^{\top}]- \mathbb{E} [   \phi(\bb{z})] \mathbb{E} [\phi(\bb{g})]^{\top} \|^2_{HS}.
    \label{eq:hsicdefine}
\end{align}
}
The norm  $\|\cdot\|_{HS}$ denotes the Hilbert-Schimidt norm, which reduces to the matrix Frobenious norm in finite dimension cases. The kernel mapping function $\phi$ maps the inlier $\bb{x}$ and outlier $\bb{a}$ into the reproducing kernel Hilbert spaces (RKHS). Hence, HSIC essentially reflects the {\emph{dependence}} between the two random variables $\bb{z}, \bb{g}$ by first taking any kernel transformation of each other, and then computing the cross covariance between these features. By using kernel trick $k(\bb{z},\bb{z}')=\langle\phi(\bb{z}),\phi(\bb{z}')  \rangle$, Eq.(\ref{eq:hsicdefine}) further reduces to:
{\small
\begin{align}
    \text{HSIC}(\bb{z},\bb{g}) & =\mathbb{E}_{z_i, z_j, g_i, g_j}[k(\bb{z}_i,\bb{z}_j)k(\bb{g}_i,\bb{g}_j)]   \nonumber                                 \\
                         & ~~~~~ -2\mathbb{E}_{z_i,g_i}[ [\mathbb{E}_{z_j}k(\bb{z}_i,\bb{z}_j)] [\mathbb{E}_{g_j}k(\bb{g}_i,\bb{g}_j)] ]\nonumber \\
                         & ~~~~~ +\mathbb{E}_{z_i, z_j}[k(\bb{z}_i,\bb{z}_j)] \mathbb{E}_{g_i, g_j}[k(\bb{g}_i,\bb{g}_j)].
    \label{eq:EEEhsic}
\end{align}
}
It manifests from Eq.(\ref{eq:EEEhsic}) that HSIC essentially measures the magnitude of the correlation in these non-linear features via mapping $\phi$. We take inspiration from this point, and we would discuss in Section \ref{sec:anal} the advantage of viewing HSIC from a correlation perspective. Such connection with non-linear correlation will eventually arrive at a novel statistical test exclusively suitable for HOOD training criteria.

\subsection{Implementation of HOOD}
HSIC can be computed efficiently. In \cite{Grettonhsic2005}, the above objective Eq.(\ref{eq:EEEhsic}) can be estimated empirically with a biased estimator, where features ($\bb{z}_1, \bb{g}_1$) ,..., ($\bb{z}_N, \bb{g}_N$) are $N$ random independent copies drawn from $p(\bb{z}, \bb{g})$, given current $\bb{\theta}$:
{\small
\begin{equation}
    {\widehat{\text{HSIC}}}(\bb{z},\bb{g})=\frac{1}{(N-1)^2}tr(\bb{K}_{z}\bb{H}\bb{K}_{g}\bb{H}).
    \label{eq:tracehsic}
\end{equation}
}
Here, the matrices are of dimensions $\bb{K}_{z}, \bb{H}, \bb{K}_{g} \in \mathbb R^{N \times N}$. For matrix $\bb{K}_z$, the $(i,j)^{th}$ entry is computed as $K_{z,i,j}=k(\bb{z}_i,\bb{z}_j)$. Similarly, for matrix $\bb{K}_g$, the $(i,j)^{th}$ entry is denoted as $K_{g,i,j}=k(\bb{g}_i,\bb{g}_j)$. $\bb{H}=\bb{I}-\frac{1}{N}\boldsymbol 1\boldsymbol 1^{\top}$, and notation $tr(\cdot)$ computes matrix trace.

In principle, the kernel function $k(\cdot,\cdot)$ can take any legitimate form of characteristic kernel in the RKHS, and any kernel will suffice to guarantee that HSIC is $0$ if and only if the random variables are independent. In the context of OOD detection, we empirically find applying radial basis function (RBF) kernel offers satisfactory performance for our purpose. We therefore by default employ the non-linear RBF kernel function to display all the empirical study throughout this paper, which is defined as:
\begin{equation}\small
    K_{z,i,j}=k(\bb{z}_i,\bb{z}_j)=\exp \left(  -\frac{\| \bb{z}_i-\bb{z}_j \|^2_2}{2\sigma^2}   \right).
\end{equation}
The kernel matrix $\bb{K}_g$ takes similar form of $\bb{K}_z$ through replacing the $\bb{z}$ vectors by OOD features $\bb{g}$. The kernel temperature $\sigma^2$ is a hyperparameter to be predefined.

During each end-to-end training iteration, we jointly sample pairs of inlier and outlier ($\bb{x}_1, \bb{a}_1$) ,..., ($\bb{x}_N, \bb{a}_N$) from the entire training data distribution to form a training batch having an effective batchsize $2N$. We then extract the features of these samples jointly observed by the deep estimator $\bb{\theta}$, and we plug the corresponding ($\bb{z}_1, \bb{g}_1$) ,..., ($\bb{z}_N, \bb{g}_N$) values into Eq.(\ref{eq:tracehsic}).

\subsection{Classification on in-distribution data}
Eq.(\ref{eq:tracehsic}) penalizes non-linear correlation between inlier features $\bb{z}$ and outlier features $\bb{g}$. In the meanwhile, the in-distribution data must also be constrained to maintain meaningful semantics so that outliers can learn the desired independence. We now define the probability $p(c|\bb{z}_i)$ to denote the predicted classification score across $c\in\{1,2...,C\}$. Since the ground truth label of $\bb{z}_i=f(\bb{x}_i, \bb{\theta})$ is $y^{in}_i$, we optimize the conventional cross entropy loss:
\begin{equation}\small
    {\mathcal{L}}_{cls}=-\frac{1}{N}\sum_i \log p(y^{in}_i| \bb{z}_i).
\end{equation}
The predication $p(y^{in}_i| \bb{z}_i)$ is obtained via a standard softmax classifier:
\begin{equation}\small
    p(y^{in}_i| \bb{z}_i)=\frac{\exp\left[ \bb{w}_{y^{in}_i}^{\top}\bb{z}_i\right]}{\sum_c \exp \left[ \bb{w}_{c}^\top\bb{z}_i\right] },
\end{equation}
where $\bb{w}_{c}$  is the classifier parameter of the $c$ class. The training then proceeds by minimizing the above classification loss with regard to both the classifier parameter $\bb{w}_{c}$ and features $\bb{z}_i$, where subscript $c$ ranges over all classes of in-distribution data.

\begin{center}
    \begin{algorithm}[t]
        \caption{HOOD: PyTorch-like Pseudocode}
        \footnotesize
        \begin{algorithmic}
            \ttfamily
            \State{\color{teal} \# f, cls:\ encoder, classier}
            \State{\color{teal} \# x, y:\ inlier data and label}
            \State{\color{teal} \# a:\ outlier data}
            \State{\color{teal} \# lambda:\ HSIC loss weight}
            \State{}
            \State{{\color{teal}\# load a mini-batch (x, y, a)}}
            \State{{\fontseries{b}\selectfont \color{magenta}for} (x, y, a) {\fontseries{b}\selectfont \color{magenta}in} loader{\color{magenta}:} }
            \State{\ \ \ \  \color{teal} \# extract feature}
            \State{\ \ \ \   z {\color{magenta}=} f(x)        {\color{teal}\# (N,d)}}
            \State{\ \ \ \   g {\color{magenta}=} f(a)        {\color{teal}\# (N,d)}}
            \State{\ \ \ \  \color{teal} \# cross entropy loss}
            \State{\ \ \ \   p {\color{magenta}=} cls(z)        {\color{teal}\# (N,C)}}
            \State{\ \ \ \   ce\_loss {\color{magenta}=} CrossEntropyLoss(p, y)  {\color{teal}\# Eq.(6)}}
            \State{\ \ \ \  \color{teal} \# hsic loss}
            \State{\ \ \ \   hsic\_loss {\color{magenta}=} HSIC(z, g)   {\color{teal}\# Eq.(4)}}
            \State{\ \ \ \  \color{teal} \# overall loss}
            \State{\ \ \ \   loss {\color{magenta}=} ce\_loss {\color{magenta}+}  lambda {\color{magenta}*} hsic\_loss {\color{teal}\# Eq.(8)} }
            \State{\ \ \ \  loss.backward()}
            \State{\ \ \ \ update(f, cls)}
            \State{}
            \State{\color{teal} \# HSIC loss}
            \State{{\fontseries{b}\selectfont \color{magenta}def} HSIC(z, g){\color{magenta}:}}
            \State{\ \ \ \ N {\color{magenta}=} z.size(0)}
            \State{\ \ \ \  \color{teal} \# compute kernel K}
            \State{\ \ \ \ K\_z {\color{magenta}=} kernel(z)    {\color{teal}\# (N,N)}}
            \State{\ \ \ \ K\_g {\color{magenta}=} kernel(g)    {\color{teal}\# (N,N)}}
            \State{\ \ \ \  \color{teal} \# compute KH}
            \State{\ \ \ \ KH\_z {\color{magenta}=} K\_z - mean(K\_z, keepdim{\color{magenta}=}True)}
            \State{\ \ \ \ KH\_g {\color{magenta}=} K\_g - mean(K\_g, keepdim{\color{magenta}=}True)}
            \State{\ \ \ \ {\fontseries{b}\selectfont \color{magenta}return} trace(mm(KH\_z, KH\_g)){\color{magenta}/}(N-1){\color{magenta}**}2}
        \end{algorithmic}
        \label{alg:HOOD}
    \end{algorithm}
\end{center}

\subsection{Overall loss function}
As per our discussions above, the overall loss function ${\mathcal{L}}_{cls}$ combines the supervision loss on inlier data and the empirical independence estimate ${\widehat{\text{HSIC}}}(\bb{z},\bb{g})$:
\begin{equation}
    \mathcal L=\mathcal L_{cls}+\lambda {\widehat{\text{HSIC}}}(\bb{z},\bb{g}). \\
    \label{eq:overall}
\end{equation}
The overall loss function then backpropagates through both the inlier features $\bb{z}_i$, the outlier feature $\bb{g}_j$, and the classifier $\bb{w}_c$ during training. Here $\lambda$ is a predefined hyperparameter balancing the strength of the desired independence. Please see Algorithm \ref{alg:HOOD} for Pseudocode of HOOD.

\subsection{HOOD without knowing $\mathcal{D}_{tr,ood}$} \label{sec:HOODun}
A much more general scenario is when $\mathcal{D}_{tr,ood}$ becomes unavailable during training. In this scenario, we simply generate fake OOD training data by applying strong distortion to inliers in the ambient space (pixel space). This solution also corresponds with precedents in literature that {\emph{extreme image distortions}} would incur strong distributional shift in features \cite{sinha2021negative,chen2021style,chen2021transferrable}. We use a simple and cheap distortion: strong augmentation \cite{cubuk2020randaugment} to generate fake OOD data ``made up'' from inliers. We compare HOOD with SOTA discriminative OOD detection methods including those only consider modeling on inliers \cite{liang2020enhancing,Mdistance2018}. Empirical results show that the basic intuitions elucidated behind HOOD still well apply even if we lose access to known real training OOD data, and HOOD leads to much superior detection performance than comparable counterparts, such as ODIN~\cite{liang2020enhancing} and Mahalanobis~\cite{Mdistance2018}.

\section{Analysis} \label{sec:anal}
\subsection{Connections with other metrics}\label{sec:connection}
We discuss the intrinsic advantage of applying the HSIC metric to the OOD detection problem. We firstly disambiguate HSIC and one seemingly applicable discrepancy metric $\text{MMD}(p(\bb{z})||p(\bb{g}))$, i.e., we compare:
\begin{equation}
\text{MMD}(p(\bb{z}, \bb{g})||p(\bb{g})p(\bb{z})) \ \   v.s. \ \   \text{MMD}(p(\bb{z})||p(\bb{g}))
\label{eq:MMDhsicvsmmd}.
\end{equation}
Optimizing the distributional dependency $\text{MMD}(p(\bb{z})||p(\bb{g}))$ for arbitrary $\bb{z}$ and $\bb{g}$ revolutionized many machine learning applications, e.g., Domain Generalization (DG) \cite{LiMMDDG2018} and Domain Adaptation (DA) \cite{long2015learning,pan2019transferrable}. One might also feel tempted to use MMD (Maximum Mean Discrepancy) penalty on OOD detection tasks. Note that minimizing the MMD discrepancy only seeks the disjoint distribution between $p(\bb{z})$ and $p(\bb{g})$, which is fundamentally different from our independence assumption. Such discrepancy also fails to reflect the discriminative OOD detection assumption, i,e., the support of class distribution $p(y^{out}_i)$ and $p(y^{in}_i)$ for all $i$ are defined as disjoint $\mathcal{Y}_{in}\cap\mathcal{Y}_{out}=\varnothing$. The metric $\text{MMD}(p(\bb{z})||p(\bb{g}))$ does not necessarily remove the feature dependence between the inlier and outlier (even not linear correlation). We should be careful not to confuse left hand side Eq.(\ref{eq:MMDhsicvsmmd}) with right hand side Eq.(\ref{eq:MMDhsicvsmmd}). Take for instance, for a variable $\bb{z}=10000-3\cdot \bb{g}$, the distribution discrepancy between the two vectors $\bb{z}$ and $\bb{g}$ can be significantly large and even are disjoint, whereas as the variable $\bb{z}$ is only a scaled version of feature $\bb{g}$ and is fully dependent on variable $\bb{g}$. Also, note that disjoint distribution between two variables does not reflect in any sense that the two variables are independent, correlated or not. Such disjoint distribution on features also does not necessarily imply that the support of outlier class distribution $p(y^{out}_i)$ and inlier $p(y^{in}_i)$, $\forall i,j$ are disjoint, i.e., $\mathcal{Y}_{in}\cap\mathcal{Y}_{out}=\varnothing$, deviating from our objective.

It is also helpful to consider the connection between the HSIC and cross-covariance matrix as discussed in \cite{tsai2021note}. Define $\bb{Z} \in \mathbb{R}^{N \times d}$ matrix as the matrix having each $\bb{z}_i^\top$ as rows, and define $\bb{G} \in \mathbb{R}^{N \times d}$ matrix as the matrix having each $\bb{g}_i^\top$ as rows. If the features $\bb{z}_i$ and $\bb{g}_i$ are both standardized vectors, while we use a linear kernel function for $k(\bb{z}_i,\bb{z}_j)$, then the empirical HSIC metric in Eq.(\ref{eq:tracehsic}) boils down to penalizing the Frobenius norm of covariance matrix $\bb{C}$:
\begin{equation}
    {\widehat{\text{HSIC}}}(\bb{z},\bb{g})=\frac{1}{(N-1)^2}tr(\bb{Z}\bb{Z}^{\top}\bb{G}\bb{G}^{\top})=  \|\bb{Z}^{\top}\bb{G}\|^2_F  = \|\bb{C}\|^2_F.
    \label{eq:covariance}
\end{equation}
Here matrix $\bb{C}=\bb{Z}^{\top}\bb{G} \in \mathbb{R}^{d \times d}$ is the conventional cross-covariance matrix that measures all cross-dimensional correlations. Under this more transparent linear kernel construction, it can be proved that HSIC not only penalizes the energy of on-diagonal entries of $\bb{C}$, which essentially measures the magnitude of linear correlation between vectors $\bb{z}_i$ and $\bb{g}_j$, it also additionally suppresses any cross-dimension correlation between $\bb{z}_i$ and $\bb{g}_j$ vectors, echoing the name of {cross-covariance matrix}.
In other words, HSIC training metric can effectively penalize an upper bound (see supplementary file for proof) of the linear correlation magnitude between two variables, and would necessarily achieve zero linear correlation if HSIC$=0$.

\subsection{Statistical test for inference}\label{sec:test}
Unfortunately, during test time, we lose access to i.i.d. data from the test OOD data distribution, and it becomes no longer feasible to compute the ${\bb{K}}_g$ matrix. It is nonetheless possible to show that the original HSIC loss can effectively penalize linear correlation in the sense that zero-valued HSIC necessarily achieves zero correlation between the two (see supplementary file for proof). This can be observed through the connection between HSIC and cross-covariance in Eq.(\ref{eq:covariance}). Such connection then inspires us to rethink the suitable test for HOOD. Since HSIC metric will necessarily also suppress the linear correlation between in and out data, then if test data shows strong linear correlations with any in-distributional classes, it incurs non-zero HSIC loss, and necessarily identifies dependence with in-distribution data (which is also common sense in statistics without our context), and is more likely to be inliers. Energy of linear correlation can then be seen as a relaxed surrogate function. Using this surrogate as the test metric, we propose:
\begin{align}
\label{eq:testhsic}
    Q_{hsic}=\max_c |\bb{\mu}_c^{\top} \bb{q}_{test}|.
\end{align}
Here, $\bb{q}_{test}=f(\bb{x}_{test}, \bb{\theta})$ extracts the feature of test data $\bb{x}_{test}$ through the obtained deep estimator $\bb{\theta}$, and:
\begin{equation}
    \bb{\mu}_c=\frac{1}{N_c}\sum_i {\mathds 1}\{y_i=c\}\bb{z}_i.
\end{equation}
The value $\bb{\mu}_c$ is the mean of feature values of class $c$ in inlier training data, where the summation counts the total number $N_c$ of training inliers in class $c$. We determine if a certain test sample is OOD data by following a thresholding policy given hyperparameter $\tau$:
\begin{equation}
    \ell(\bb{x}_{test}, \tau, \bb{\theta})=
    \begin{cases}
        1,& \text{if } Q_{hsic}\le\tau,\\
        0,              & \text{otherwise,}
    \end{cases}
\end{equation}
where $\ell(\bb{x}_{test}, \tau, \bb{\theta})=1$ classifies $\bb{x}_{test}$ data into OOD category, and $\ell(\bb{x}_{test}, \tau, \bb{\theta})=0$ classifies $\bb{x}_{test}$ into inlier category. We compute the class-wise correlation between each test sample and each class, as any non-zero correlation with the subset of inlier data will necessarily identify the non-zero dependence.

HOOD can achieve good generalization because HSIC loss poses a strict constraint during training. This constraint requires that OOD data can only span their subspace using independent features than inliers. This tighter training constraint is therefore also valid and beneficial when applied with other related tests. In the empirical study, we show that a simple softmax test on the HOOD trained feature also can beat the SOTA methods by a large margin.

\section{Experiments}
In this section, we evaluate the efficiency of HOOD and the associated novel statistical test to justify our proposed hypothesis. We consider assessing all the methods with three evaluation metrics: area under the receiver operating characteristic curve (AUROC), area under the prevision-recall curve (AUPR), and the false positive rate at $95\%$ true positive rate (FPR95). These metrics well reflect the separability of inliers and outliers by ranging the statistical test across different threshold values. For a more detailed interpretation of these metrics, please refer to \cite{hendrycks2018deep}.

\subsection{Baseline methods}
We compare HOOD with several important baselines. The approaches under consideration fall into two categories regarding their training strategies, although they all follow discriminative model evaluation protocols and definitions $\mathcal{Y}_{in}\cap\mathcal{Y}_{out}=\varnothing$. The first category requires training with OOD data, including the state-of-the-art methods: OE \cite{hendrycks2018deep}, Energy \cite{Liuenergy2020}, MMD baseline. The MMD baseline corresponds to the training loss using Eq.(\ref{eq:MMDhsicvsmmd}). The second category requires no OOD data during training, and these baseline models mainly focus on the effect of the statistical test during inference. Such methods include: MSP~\cite{hendrycks17baseline}, ODIN \cite{liang2020enhancing}, Mahalanobis-distance \cite{Mdistance2018}, G-ODIN \cite{GeneralizedODIN}, and ReAct~\cite{ReActSun2021} baseline. For the MSP baseline model, the network is trained merely on the annotated in-distributional data, and we threshold the maximum softmax prediction among classes for each test sample as in \cite{hendrycks17baseline,hendrycks2018deep}. We do not compare with further generative models here, since generative models have completely different evaluation protocols and the OOD definition. For each baseline, we download the official codes from the website, and we re-implemented all of the methods in a head-to-head fair comparison by faithfully tuning each method to our best.

\subsection{Datasets}
Our training and testing dataset are defined as follows. For training method with OOD data exposure (OE \cite{hendrycks2018deep}, Energy \cite{Liuenergy2020} and HOOD), the training dataset is a mixture of both labeled in-distribution data $\mathcal{D}_{tr,in}$ and unlabeled out-of-distribution data $\mathcal{D}_{tr,ood}$. During inference phase, we test on a mixture of in-distribution data $\mathcal{D}_{tst,in}$ and OOD data $\mathcal{D}_{tst,ood}$ with the goal to detect $\mathcal{D}_{tst,ood}$. Note that the training data preprocessing procedure \cite{hendrycks2018deep}, Energy \cite{Liuenergy2020}  ensures that there is no overlap between training OOD data $\mathcal{D}_{tr,ood}$ and test OOD data $\mathcal{D}_{tst,ood}$, i.e., $\mathcal{D}_{tr,ood} \cap\mathcal{D}_{tst,ood}=\varnothing$.

{\bf In-distribution datasets}. By following the dataset protocol defined in \cite{hendrycks2018deep}, we consider the following datasets as in-distribution datasets $\mathcal{D}_{tr,in}$: {\bf 1)} {{\bf{CIFAR-100}} \cite{Krizhevsky09learningmultiple} consists of tiny color images in 100 classes.  {\bf 2)} {\bf Place365} \cite{zhou2017places} is a large dataset for real-world scene recognition. {\bf 3)} {\bf Tiny-ImageNet}~\cite{Le2015TinyIV} is a subset sampled from ImageNet1K, in which each image is of $64\times64$ resolution.

{\bf OOD training datasets}. According to \cite{hendrycks2018deep}, we select the following dataset as the training OOD data, i.e., $\mathcal{D}_{tr,ood}$: {\bf 1)}. {\bf{80 Million Tiny Images}} \cite{torralba200880}. As suggested by \cite{hendrycks2018deep}, we remove from 80 Million Tiny Images the subsets overlapped with CIFAR datasets. {\bf 2)}. {\bf{ImageNet-21K}} \cite{ILSVRC15}. We adopt the ImageNet dataset including approximately 22 thousand classes, and we remove ImageNet1K from the ImageNet22K training data as in \cite{hendrycks2018deep}. For training with CIFAR-100 as $\mathcal{D}_{tr,in}$, we adopt 80 Million Tiny Images as $\mathcal{D}_{tr,ood}$. When Place365 or Tiny-ImageNet are used as $\mathcal{D}_{tr,in}$, we employ ImageNet-21K (with removal of ImageNet-1K) as $\mathcal{D}_{tr,ood}$.

{\bf OOD test datasets}. The OOD data used during test are: {\bf{DTD}} \cite{cimpoi14describing}, {\bf{SVHN}} ({\em test} set) \cite{shvn11}, {\bf{LSUN}} ({\em test} set) \cite{yu2015lsun}, {\textbf{Place69}} \cite{zhou2017places}, \textbf{ImageNet-1K} ({\em val} set)  \cite{ILSVRC15}, \textbf{ImageNet-800} ({\em val} set) \cite{ILSVRC15} and \textbf{CIFAR-10} ({\em test} set)   \cite{Krizhevsky09learningmultiple}. In detail, {\bf DTD} consists of textural images in the wild; {\bf SVHN} is a digit classification benchmark dataset including RGB images of printed digits cropped from pictures of house number plates;  {\bf Place69} contains 69 scene categories excluded from {\bf Place365}; {\bf LSUN} is another scene recognition dataset; {\bf CIFAR-10} includes tiny color images in 10 classes for object recognition; {\bf ImageNet1K} is a huge and broadly recognized object recognition benchmark, while {\bf ImageNet800} removes 200 classes of {\bf Tiny-ImageNet} from ImageNet1K.

\begin{table}[t]
\caption{\small Averaged~OOD~detection~performance. $\mathcal{D}_{tr,in}$ = CIFAR-100, $\mathcal{D}_{tr,ood}$ = 80 Million Tiny Images, $\mathcal{D}_{tst,ood}$ = \{DTD, SVHN, Place365, LSUN, CIFAR-10\}.}

\begin{adjustbox}{width=0.49\textwidth}\setlength\tabcolsep{3pt}
    \begin{tabular}{clcccc}
            \shline
            $\mathcal{D}_{tr,in}$   & \textbf{Methods}  & \textbf{Training OOD} & \textbf{FPR95(\%)$\downarrow$} & \textbf{AUROC(\%)$\uparrow$} & \textbf{AUPR(\%)$\uparrow$} \\ \hline
            \multirow{14}{*}{\rotatebox[origin=c]{90}{\begin{tabular}[c]{@{}c@{}}\textbf{CIFAR-100}\\ (WideResNet)\end{tabular}}}
                                                & MSP \cite{hendrycks17baseline} & \XSolidBrush                                       & 65.84                         & 74.76                       & 33.25                       \\
                                                & ODIN \cite{liang2020enhancing}  & \XSolidBrush                                          & 65.82                         & 77.08                       & 36.12                       \\
                                                & Mahalanobis \cite{Mdistance2018}  & \XSolidBrush                                      & 61.11                          & 79.12                        & 44.39                       \\
                                                & ReAct \cite{ReActSun2021}  & \XSolidBrush                                          & 59.37                         & 79.90                       & 36.62                       \\
                                                & G-ODIN \cite{GeneralizedODIN}  & \XSolidBrush                                          & 56.52                         & 80.38                       & 49.54                      \\
                                                & OE+fake  & \XSolidBrush                                                                           & 58.69                         & 80.63                        & 48.71                       \\
                                                & HOOD+fake  & \XSolidBrush                                                                     & \textbf{47.38}             & \textbf{84.07}           & \textbf{62.13}           \\ \cline{2-6}
                                                & MMD   & \Checkmark                                                                                  & 61.38                          & 80.84                        & 49.43                     \\
                                                & Energy \cite{Liuenergy2020}  & \Checkmark                                              & 43.53                          & 86.02                        & 56.94                      \\
                                                & OE\cite{hendrycks2018deep}  & \Checkmark                                             & 47.61                          & 86.82                        & 58.48                     \\
                                                & OE+aug & \Checkmark                                                                               & 58.24                          & 81.29                        & 52.17                     \\
                                                & HOOD  & \Checkmark                                                                                 & 43.81                          & 87.93                        & 59.90                     \\
                                                & HOOD+aug   & \Checkmark                                                                        & \textbf{34.44}              & \textbf{88.37}            & \textbf{67.90}        \\ \shline
        \end{tabular}
        \label{tab:cifar100}
\end{adjustbox}
\end{table}

\subsection{Training setup} \label{sec:trainsetup}
The following training setup and training policy apply to all the compared methods. For OOD detection baselines and methods under consideration in this paper, we implement all the experiments by training from scratch (with random network initialization) without any finetuning. This setup helps better differentiate the pros and cons of various OOD detection approaches, and also eases sanity checks for each method. We train all methods for 100 epochs with regard to the size of in-distributional training data $N_{tr, in}$ of a specific dataset. We define the ratio of $R_{o:i}=\frac{N_{tr,ood}}{N_{tr,in}}$, and correspondingly train with proper OOD training data size $N_{tr,ood}$ according to this $R_{o:i}$. According to \cite{hendrycks2018deep}, we use $R_{o:i}=2$ throughout this paper. We specify the network architecture used for each experiment in the corresponding evaluation tables. The employed networks include ResNet18 \cite{he2016deep} and WideResNet-40-2 \cite{Zagoruyko2016WRN}. The hyperparameter selected for each method is not tuned w.r.t. different OOD test datasets. Instead, we choose consistent hyperparameters for each specific training data pair $\mathcal{D}_{tr,ood}-\mathcal{D}_{tr,in}$, e.g., 80 Million Tiny Images $-$ CIFAR-100.

\textbf{Data augmentation.} For all plain models (e.g., the vanilla OE, Energy, HOOD), we apply the same standard augmentations on both training inliers and training outliers as suggested in \cite{hendrycks2018deep}. These augmentations include random-crop and random-flip. For all ``X+aug'' methods where ``X'' refers to the corresponding plain model, we further apply additional strong augmentation on the training OOD data. These augmentation actions include adding RandAugment~\cite{cubuk2020randaugment} to the training OOD data. Specifically, for Table \ref{tab:cifar100}, \ref{tab:places}, \ref{tab:tinyimage}, we set the magnitude of augmentation $M=10$ and the number of augmentation transformations $N=4$ for RandAugment. For all ``X+fake'' results in Table \ref{tab:cifar100}, \ref{tab:places}, \ref{tab:tinyimage}, we adopt the optimal $N=5$ and fixed $M=30$ for RandAugment to generate fake OOD data ``made up'' from inliers. In regard to the influence of strong augmentations, we include more discussions in Section ~\ref{sec:aug}.

\begin{table}[t]
\caption{\small Averaged~OOD~detection~performance. $ \mathcal{D}_{tr,in}$ = Place365,  $\mathcal{D}_{tr,ood}$ = ImageNet-21K, $\mathcal{D}_{tst,ood}$ = \{DTD, SVHN, Place69, ImageNet-1K\}.}
\begin{adjustbox}{width=0.49\textwidth}\setlength\tabcolsep{3pt}
\begin{tabular}{clcccc}
        \shline
        $\mathcal{D}_{tr,in}$   & \textbf{Methods}  & \textbf{Training OOD} & \textbf{FPR95(\%)$\downarrow$} & \textbf{AUROC(\%)$\uparrow$} & \textbf{AUPR(\%)$\uparrow$} \\ \hline
        \multirow{14}{*}{\rotatebox[origin=c]{90}{\begin{tabular}[c]{@{}c@{}}\textbf{Place365}\\ (ResNet18)\end{tabular}}}
                                                         & MSP   \cite{hendrycks17baseline}& \XSolidBrush                                      & 84.81                        & 62.40                       & 23.99                       \\
                                                         & ODIN \cite{liang2020enhancing}& \XSolidBrush                                            & 68.89                        & 73.50                        & 39.37                      \\
                                                         & Mahalanobis \cite{Mdistance2018}& \XSolidBrush                                         & 66.20                       & 75.45                        & 42.59                      \\
                                                & ReAct \cite{ReActSun2021}  & \XSolidBrush                                          & 68.07                        & 73.83                       & 39.38                        \\
                                                & G-ODIN \cite{GeneralizedODIN}  & \XSolidBrush                                          & 68.76                         & 72.96                       & 36.26                       \\
                                                         & OE+fake & \XSolidBrush                                                                              & 79.91                       & 68.86                        & 35.92                      \\
                                                         & HOOD+fake& \XSolidBrush                                                                        & \textbf{62.66}           & \textbf{76.31}            & \textbf{45.76}                                                                                               \\ \cline{2-6}
                                                         & MMD  & \Checkmark                                                                                     & 86.02                       & 63.00                        & 25.42                       \\
                                                         & Energy \cite{Liuenergy2020} & \Checkmark                                                 & 58.27                       & 80.29                        & 51.29                       \\
                                                         & OE \cite{hendrycks2018deep}  & \Checkmark                                              & 78.91                       & 73.09                        & 38.46                       \\
                                                         & OE+aug   & \Checkmark                                                                               & 77.73                       & 71.53                        & 41.83                       \\
                                                         & HOOD     & \Checkmark                                                                               & 53.85                        & 81.15                       & 48.57                        \\
                                                         & HOOD+aug & \Checkmark                                                                           & \textbf{48.29}            & \textbf{83.51}          & \textbf{58.54}            \\ \shline
    \end{tabular}
    \label{tab:places}
    \end{adjustbox}
\end{table}

\textbf{Optimization.} We train all the models from scratch for 100 epochs w.r.t. the size of inlier training data, and we follow the default training pipeline as described in \cite{hendrycks2018deep}. For models trained with CIFAR-100 and Tiny-ImageNet as training inliers, we employ the WideResNet-40-2~\cite{Zagoruyko2016WRN} with a dropout rate of $0.3$. And we adopt an SGD optimizer with a base learning rate of $lr=0.1$, Nesterov momentum of $0.9$, and weight decay of $5\times10^{-4}$. The learning rate follows a cosine learning rate decay schedule~\cite{loshchilov2016sgdr} with a ``final learning rate'' of $10^{-5}$. The effective training batch size is $N_{tr,in}+N_{tr,ood}$, where $N_{tr,in}=128$ for inlier data and $N_{tr,ood}=256$ for outlier data. For Place365, we adopt the ResNet18~\cite{he2016deep} as our backbone, and we use an initial learning rate of $lr=0.2$ and weight decay of $1\times10^{-4}$. These models are trained with $N_{tr,in}=256$ inlier samples and $N_{tr,ood}=512$ outlier samples during each training iteration. Particularly, during HOOD's each training iteration, we use the first half of the OOD data batch, i.e., of size $N_{tr,ood}/2$ to plug into our training loss Eq.(\ref{eq:tracehsic}) associating with each $N_{tr,in}$ size of training inliers. We then use the remaining half of the OOD data batch of size $N_{tr,ood}/2$ to pair with the $\emph{same}$ $N_{tr,in}$ training inliers, and we average these two $\widehat{\text{HSIC}}(\bb{z},\bb{g})$ estimates obtained via Eq.(\ref{eq:tracehsic}) during each training iteration.

\begin{table}[t]
\caption{\small Averaged~OOD~detection~performance. $\mathcal{D}_{tr,in}$ = Tiny-ImageNet,  $\mathcal{D}_{tr,ood}$ = ImageNet-21K, $\mathcal{D}_{tst,ood}$ = \{DTD, SVHN, Place365, LSUN, ImageNet-800\}.}
\begin{adjustbox}{width=0.49\textwidth}\setlength\tabcolsep{3pt}
    \begin{tabular}{clcccc}
        \shline
        $\mathcal{D}_{tr,in}$   & \textbf{Methods}  & \textbf{Training OOD} & \textbf{FPR95(\%)$\downarrow$} & \textbf{AUROC(\%)$\uparrow$} & \textbf{AUPR(\%)$\uparrow$} \\ \hline
        \multirow{14}{*}{\rotatebox[origin=c]{90}{\begin{tabular}[c]{@{}c@{}}\textbf{Tiny-ImageNet}\\ (WideResNet)\end{tabular}}}
         & MSP     \cite{hendrycks17baseline}  & \XSolidBrush                                   & 62.83                       & 76.83                        & 39.53                       \\
                                                         & ODIN \cite{liang2020enhancing}  & \XSolidBrush                                           & 59.69                       & 78.07                        & 41.77                       \\                                                         & Mahalanobis \cite{Mdistance2018}  & \XSolidBrush                                        & 79.69                       & 61.47                        & 25.57                       \\
                                                & ReAct \cite{ReActSun2021}  & \XSolidBrush                                          & 60.51                         & 77.66                       & 42.12                      \\
                                                & G-ODIN \cite{GeneralizedODIN}  & \XSolidBrush                                          & 45.01                         & 84.31                       & 55.38                       \\

                                                         & OE+fake   & \XSolidBrush                                                                            &  19.19                      &  96.56                       &  90.51                      \\
                                                         & HOOD+fake  & \XSolidBrush                                                                       & \textbf{1.70}             & \textbf{99.48}            & \textbf{96.38}           \\ \cline{2-6}

                                                         & MMD   & \Checkmark                                                                                      & 0.77                         & 99.37                        & 93.17                        \\
                                                         & Energy  \cite{Liuenergy2020}   & \Checkmark                                                & \textbf{0.00}             & 99.94                        & 99.89                        \\
                                                         & OE  \cite{hendrycks2018deep}   & \Checkmark                                              & 7.77                         & 98.54                        & 96.62                        \\
                                                         & OE+aug    & \Checkmark                                                                                & 11.23                        & 97.34                        & 93.18                        \\
                                                         & HOOD     & \Checkmark                                                                                 & 0.01                          & 99.98                        & 99.77                        \\
                                                         & HOOD+aug    & \Checkmark                                                                          & \textbf{0.00}              & \textbf{99.99}           & \textbf{99.97}            \\         \shline
    \end{tabular}
    \label{tab:tinyimage}
\end{adjustbox}
\end{table}

\textbf{OOD data sampling policy.} As mentioned in the main paper, we have removed the randomness from OOD training samples by seeding the OOD data. This is because the sizes of the complete 80 Million Tiny Images ~\cite{torralba200880} and ImageNet-21K~\cite{ILSVRC15} are huge. Sampling a small fraction of data from these OOD training data will inevitably introduce evaluation randomness for each method. This randomness has caused comparison fairness issues, since different methods are actually trained on different OOD training data. We therefore sample a fixed subset of data from $\mathcal{D}_{tr,ood}$ to eliminate the data randomness. We sample in total $n$ groups of OOD samples to associate with the $n$ inline training epochs. In detail, by following \cite{hendrycks2018deep}, during each training epoch, 1 out of $n$ group of training OOD data is used to associate with the inlier's training for a single epoch. By doing so, all methods will be exposed to the same subset OOD training data even across different training runs. In addition, these distinct data groups are selected by uniformly sampling from $\mathcal{D}_{tr,ood}$. To evaluate the impact of different sampling policies, we particularly compare our used {\em fixed} policy described above versus the {\em random} sampling method as used in \cite{hendrycks2018deep} with the training data pair $\text{CIFAR-100} - \text{80 Million Tiny Images}$. As shown in Table~\ref{tab:diff_sample}, these two sampling strategies (i.e., {\em random} and {\em fixed}) perform comparably showing very closed results, whereas we have effectively eliminated the undesired training data randomness via {\em fixed} policy.

\subsection{Comparisons with state-of-the-art}
In this section, we compare different OOD detection methods respectively under the FPR95, AUROC, and AUPR metrics. For approaches without OOD training data exposures, i.e., MSP, ODIN, Mahalanobis, ReAct, and G-ODIN method, the models reported in Table \ref{tab:cifar100} are only trained on $\mathcal{D}_{tr,in}$ = CIFAR-100. For all methods with OOD data exposures, i.e., MMD, Energy, OE, and HOOD, the models are $ \mathcal{D}_{tr,in}$ = CIFAR-100 along with $\mathcal{D}_{tr,ood}$ = 80 Million Tiny Images,  the corresponding models are tested respectively on dataset $\mathcal{D}_{tst,ood}$ = \{DTD, SVHN, Place365, LSUN, CIFAR-10\} with in-distributional test data $\mathcal{D}_{tst,in}$ = CIFAR-100. Following a similar evaluation policy, in Table \ref{tab:places}, the methods with OOD data exposures are trained on both $ \mathcal{D}_{tr,in}$ = Place365 along with $\mathcal{D}_{tr,ood}$ = ImageNet-21K, which are then tested on $\mathcal{D}_{tst,ood}$ = \{DTD, SVHN, Place69, ImageNet-1K\} with in-distributional test data $\mathcal{D}_{tst,in}$ = Place365. In Table \ref{tab:tinyimage}, the methods with OOD data exposures are trained on both $ \mathcal{D}_{tr,in}$ = Tiny-ImageNet and $\mathcal{D}_{tr,ood}$ = ImageNet-21K, which are then tested on $\mathcal{D}_{tst,ood}$ = \{DTD, SVHN, Place365, LSUN, ImageNet-800 (ImageNet1K without Tiny-ImageNet)\}. Following \cite{hendrycks2018deep}, the ratio of ${N}_{tst,in}:{N}_{tst,out}=5:1$ in all experiments.

According to~\cite{hendrycks2018deep}, we adopt the same image pre-process procedure for all test images. We firstly fix the shorter side of the image to $S$ and keep the image aspect ratio. We then crop $S \times S$ center region for each test image. Specifically, we set $S=32$ in the $\mathcal{D}_{tr,in}=\text{CIFAR-100}$ experiment, while we set $S=64$ for the $\mathcal{D}_{tr,in}=\text{\{Place365 or Tiny-ImageNet\}}$ experiments.

\begin{table}[t]
	\begin{center}
	\caption{\small Ablation study on different OOD data sampling policies.}
	\begin{adjustbox}{width=0.49\textwidth}\setlength\tabcolsep{6pt}
        \begin{tabular}{lccc}
        \shline
        \textbf{Method}                                                         & \textbf{FPR95(\%)$\downarrow$} & \textbf{AUROC(\%)$\uparrow$} & \textbf{AUPR(\%)$\uparrow$} \\ \hline
         OE~\cite{hendrycks2018deep} ({\em random}) & 47.05 & 86.60 & 58.83   \\
         OE~\cite{hendrycks2018deep}   ({\em fixed})                      & 47.61     & 86.82     & 58.48                 \\
         HOOD  ({\em random})                                                        & 34.73          & 88.14          & 67.25                       \\
         HOOD  ({\em fixed})                                                             & 34.44     & 88.37     & 67.90                  \\
        \shline
        \end{tabular}
        \label{tab:diff_sample}
        \end{adjustbox}
        \end{center}
\end{table}

Every table reports the mean values by averaging across the total number of test datasets. Each specific training setup was trained for 5 different runs, and each trained model out of a single training run was tested 10 times. The mean value of each metric is reported in the tables. Performances on each specific test dataset are provided in the supplementary material.

\begin{figure*}[t]
    \centering
    \begin{subfigure}{0.325\textwidth}
        \includegraphics[width=\textwidth]{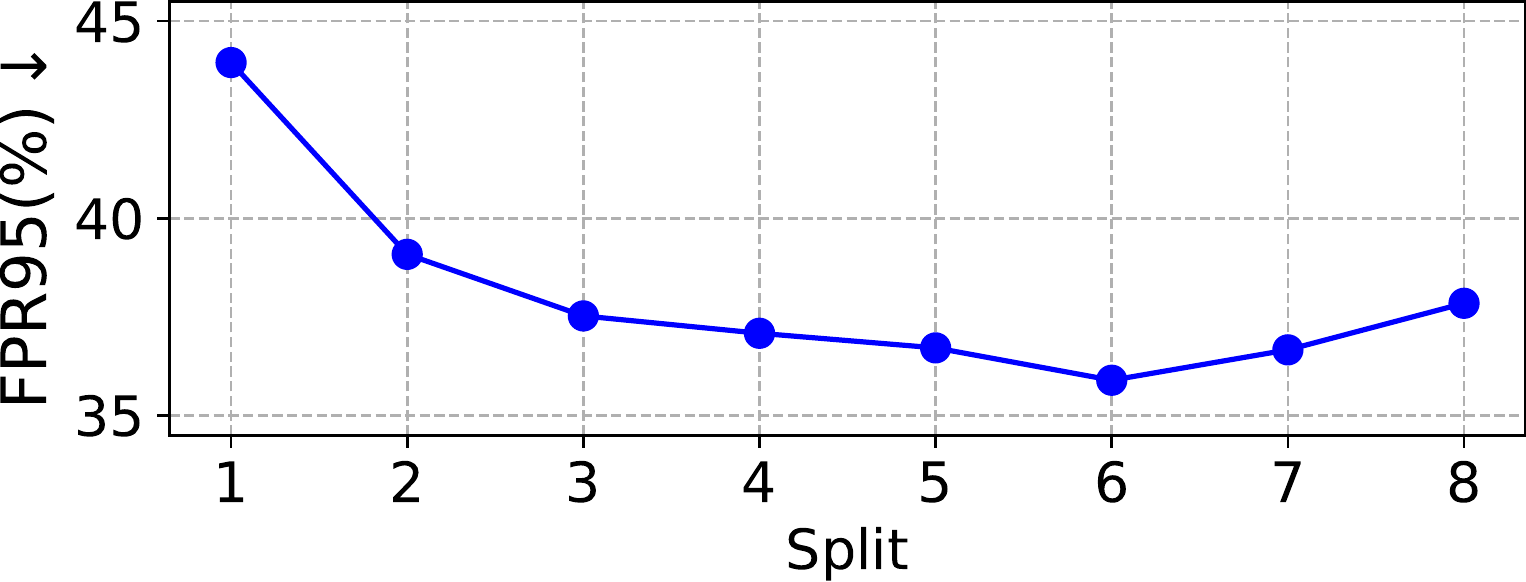}
        \caption{FPR95 v.s. Training OOD difficulty}
        \label{fig:difficulty1}
    \end{subfigure}
    \begin{subfigure}{0.325\textwidth}
        \includegraphics[width=\textwidth]{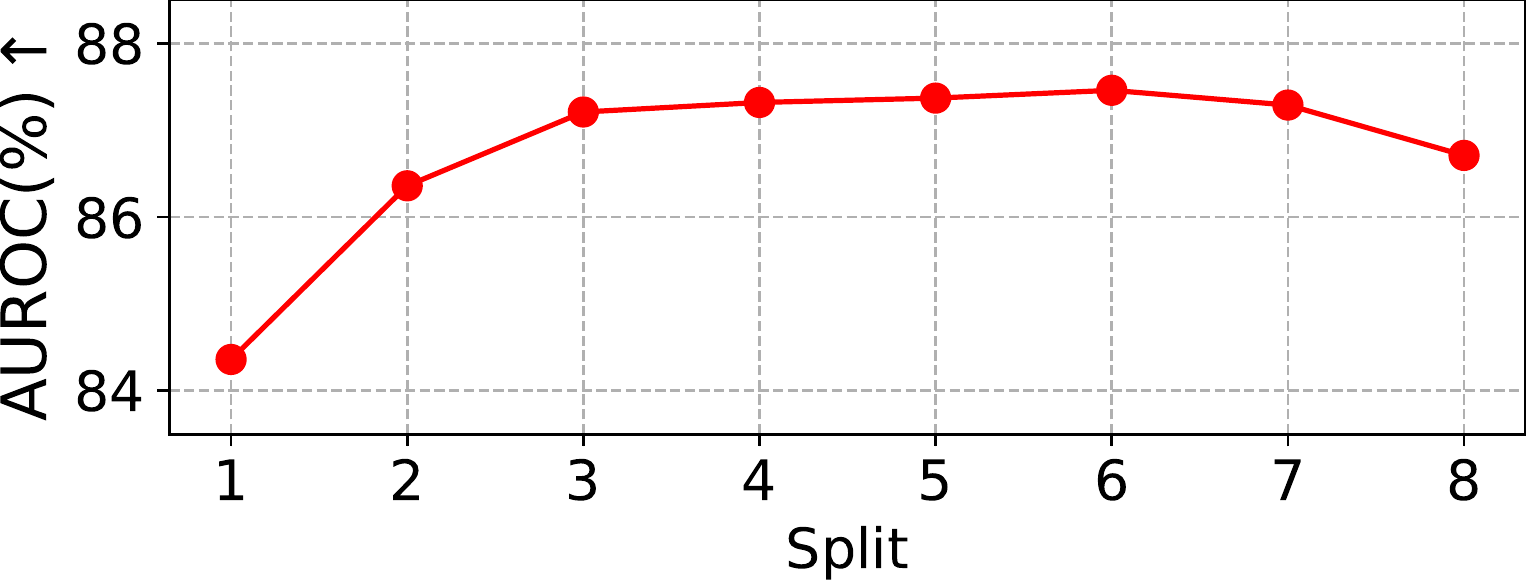}
        \caption{AUROC v.s. Training OOD difficulty}
        \label{fig:difficulty2}
    \end{subfigure}
    \begin{subfigure}{0.325\textwidth}
        \includegraphics[width=\textwidth]{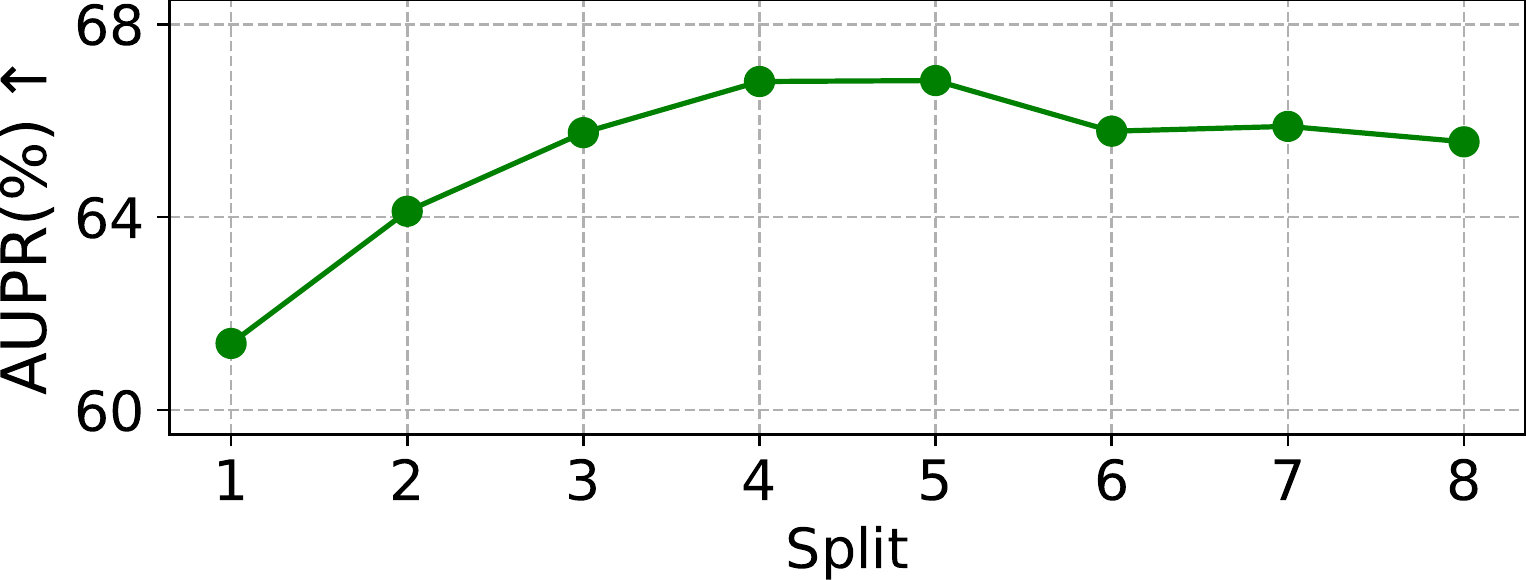}
        \caption{AUPR v.s. Training OOD difficulty}
        \label{fig:difficulty3}
    \end{subfigure}
    \\ \vspace{1em}
    \begin{subfigure}{0.325\textwidth}
        \includegraphics[width=\textwidth]{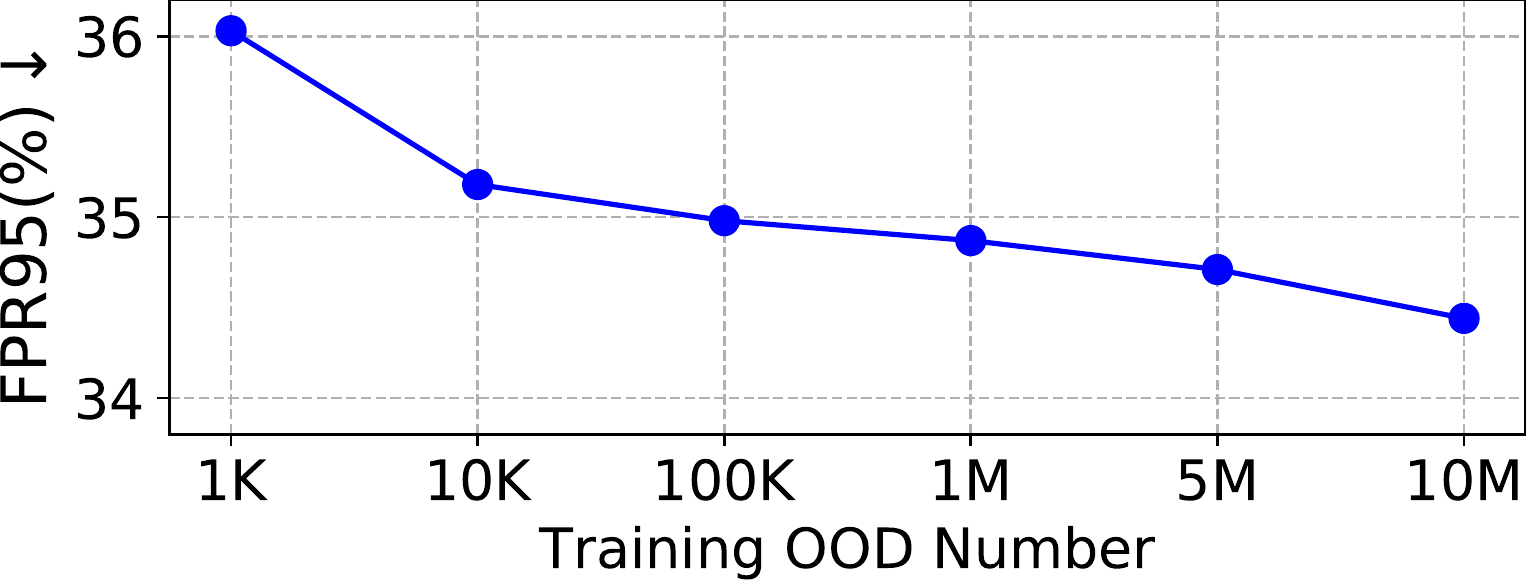}
        \caption{FPR95 v.s. Training OOD number}
        \label{fig:size1}
    \end{subfigure}
    \begin{subfigure}{0.325\textwidth}
        \includegraphics[width=\textwidth]{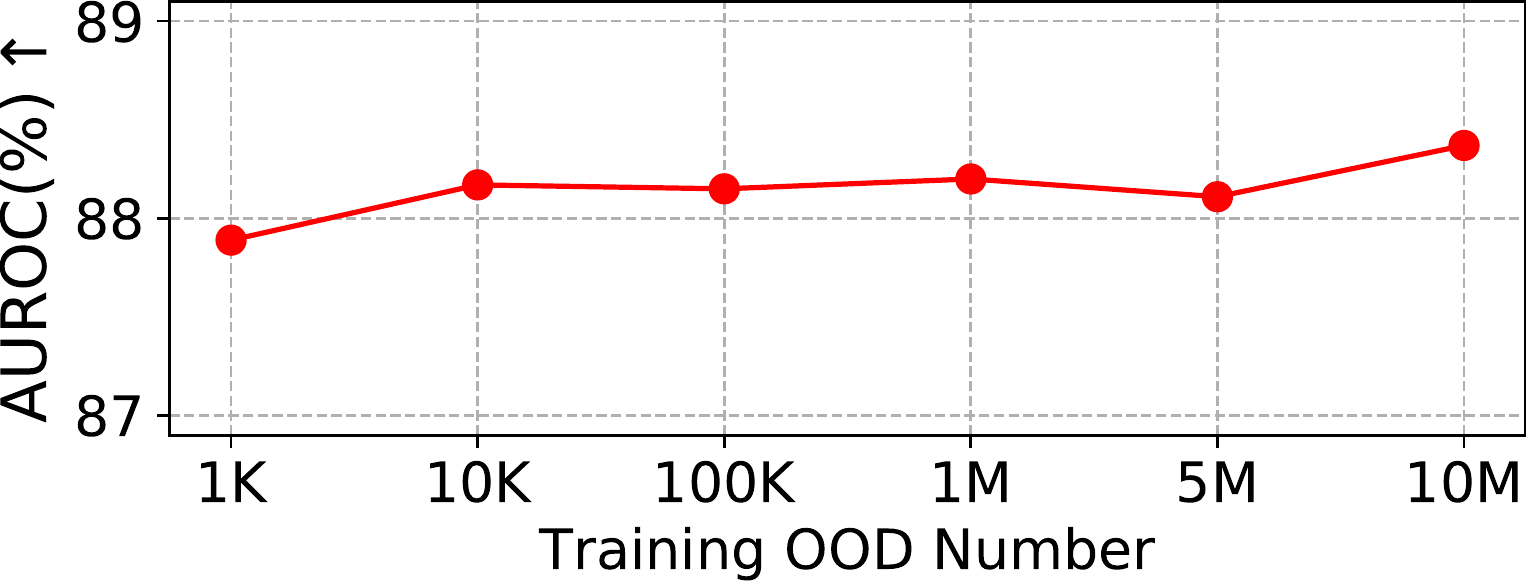}
        \caption{AUROC v.s. Training OOD number}
        \label{fig:size2}
    \end{subfigure}
    \begin{subfigure}{0.325\textwidth}
        \includegraphics[width=\textwidth]{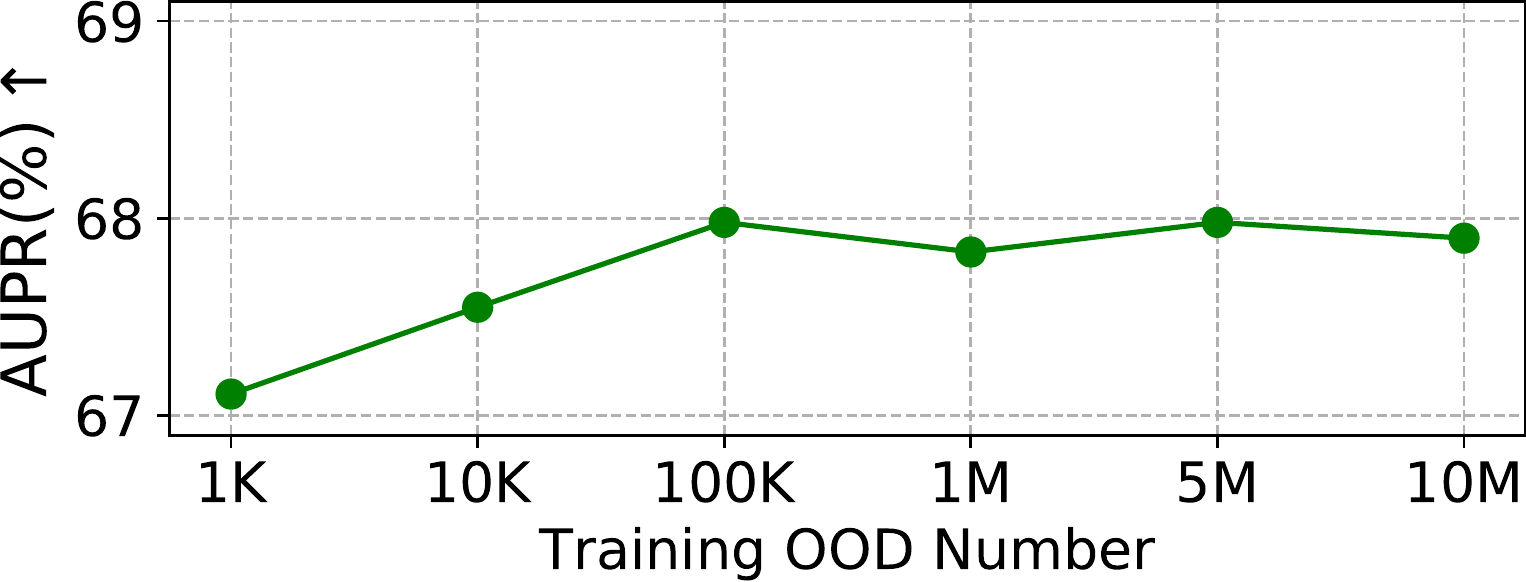}
        \caption{AUPR v.s. Training OOD number}
        \label{fig:size3}
    \end{subfigure}
    \caption{Ablation studies against the impact of training OOD data distribution and the size of training OOD data.}
    \label{fig:abaltionall}
\end{figure*}

As Table \ref{tab:cifar100}, \ref{tab:places}, \ref{tab:tinyimage} display, our proposed HOOD method always performs the best among all compared methods across all the training datasets and test sets. By incorporating the HSIC metric, HOOD training has learned features better characterizing the OOD data than others. Also, we note that our reported OE performance trained on the CIFAR-100 benchmark (i.e., Table \ref{tab:cifar100}) dataset pretty much resembles that reported in the OE paper \cite{hendrycks2018deep}. Even if we have used a fixed group of OOD training data sampled from the 80 Million Tiny Images by replacing the random sampling strategy adopted in \cite{hendrycks2018deep}, our reproduced performance of OE is very close to the published result in \cite{hendrycks2018deep}.

Another phenomenon is that methods with OOD exposure, i.e., OE, Energy, HOOD usually perform better than the methods without access to OOD data. This shows that the OOD data during training more or less can indeed help the deep parameters learn useful differentiating information. Here, MSP method corresponds to the test method where we simply threshold the maximum softmax prediction value among classes of each test sample. As seen from Table \ref{tab:cifar100}, ODIN, Mahalanobis, ReAct, and G-ODIN methods are all superior to the plain form of softmax prediction used for MSP, showing some intrinsic drawbacks of the plain form softmax prediction metric.

The MMD baseline corresponds to the training loss left hand side Eq.(\ref{eq:MMDhsicvsmmd}). Table \ref{tab:cifar100}, \ref{tab:places}, \ref{tab:tinyimage} report that, the MMD baseline that aims to separate the inliers and outliers during training can marginally beat the MSP method. It demonstrates that the network parameter does not learn many useful discriminative features through merely forcing inliers and outliers to lie apart in disjoint probability regions, as such discrimination ability does not generalize well to unseen test OOD data. In this regard, HOOD shows entirely different generalization behavior from MMD, showing clear superiority of the independence assumption.

In Table \ref{tab:cifar100}, \ref{tab:places}, \ref{tab:tinyimage}, we use ``X+aug'' to denote method ``X'' trained by applying strong augmentations \cite{cubuk2020randaugment} on OOD training data (see Section \ref{sec:trainsetup} for details). It is worthy to note that HSIC metric consistently benefits from such expanded distribution of the training OOD data in regard to FPR95 and AUPR metrics, with marginal interference on AUROC. While the OE baseline struggles in extracting useful information from the augmentations and performs worse than its plain form, HOOD method exclusively enjoys benefit from extra data augmentations and clearly outperforms all methods. This again illustrates HOOD's unique learning capability in differentiating OOD data. In detail, we specify the kernel temperature $\sigma$ and the HSIC weight $\lambda$ of HOOD used for producing Table \ref{tab:cifar100}, \ref{tab:places}, \ref{tab:tinyimage} as follows: for Table~\ref{tab:cifar100}, $\sigma=5.0$, $\lambda=1.0$; for Table~\ref{tab:places}, $\sigma=8.0$, $\lambda=3.0$; for Table~\ref{tab:tinyimage}, $\sigma=4.0$, $\lambda=0.6$. In terms of the influence of these two hyperparameters, we include more discussions in Section~\ref{sec:hyperparam}.

HOOD can also be well applied to scenarios when $\mathcal{D}_{tr,ood}$ becomes unavailable. We use ``HOOD+fake'' to represent the HOOD trained with fake OOD training data as described in Section \ref{sec:HOODun}. As reported in Table \ref{tab:cifar100}, \ref{tab:places}, \ref{tab:tinyimage}, HOOD+fake significantly surpasses all baselines that do not take into account OOD training data, including ODIN and Mahalanobis methods. In Table  \ref{tab:places}, \ref{tab:tinyimage}, HOOD+fake even outperforms OE with full exposure to $\mathcal{D}_{tr,ood}$, and is also much better than ``OE+fake'' trained with the same fake OOD training data.

\begin{figure*}[t]
    \centering
    \begin{subfigure}{0.325\textwidth}
        \includegraphics[width=\textwidth]{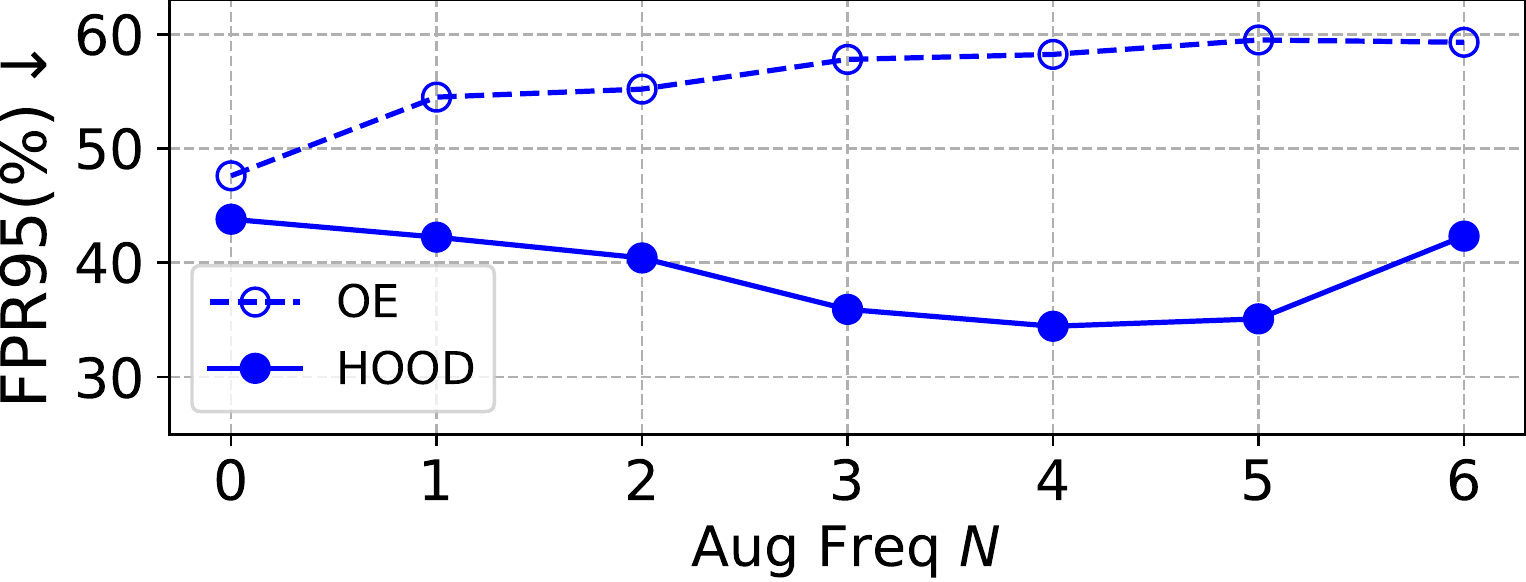}
        \caption{FPR95 v.s. RandAugment frequency $N$}
        \label{fig:augn1}
    \end{subfigure}
    \begin{subfigure}{0.325\textwidth}
        \includegraphics[width=\textwidth]{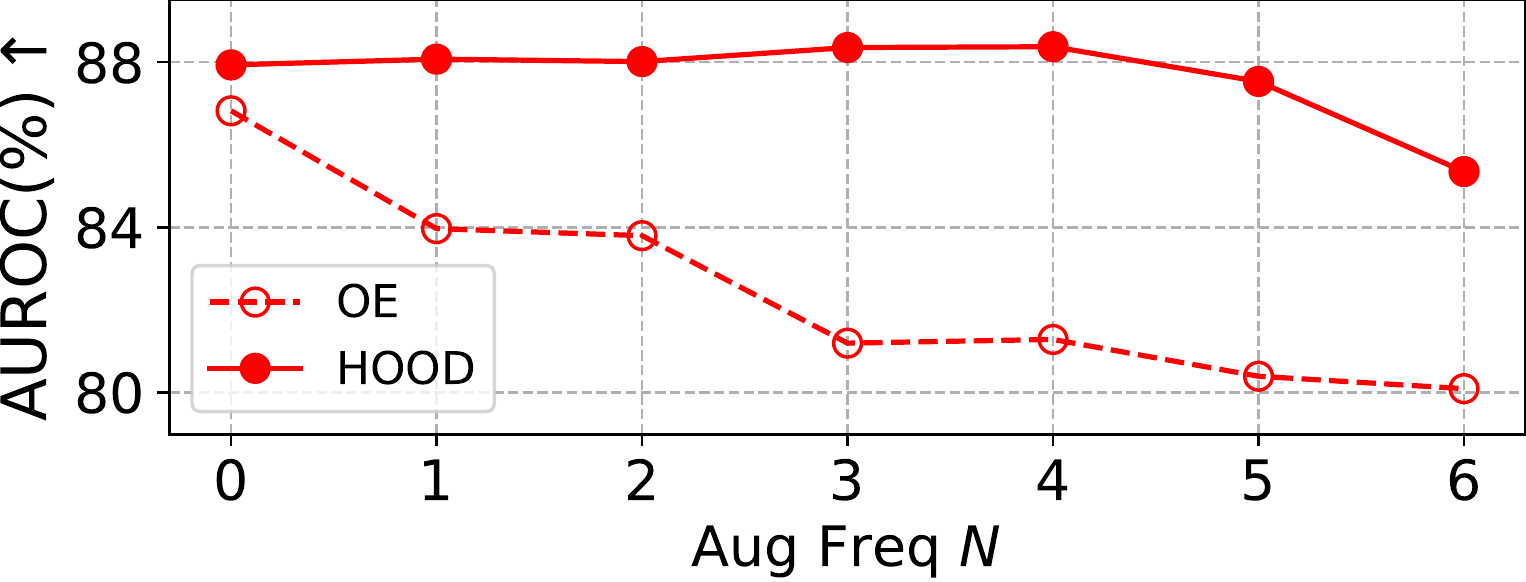}
        \caption{AUROC v.s. RandAugment frequency $N$}
        \label{fig:augn2}
    \end{subfigure}
    \begin{subfigure}{0.325\textwidth}
        \includegraphics[width=\textwidth]{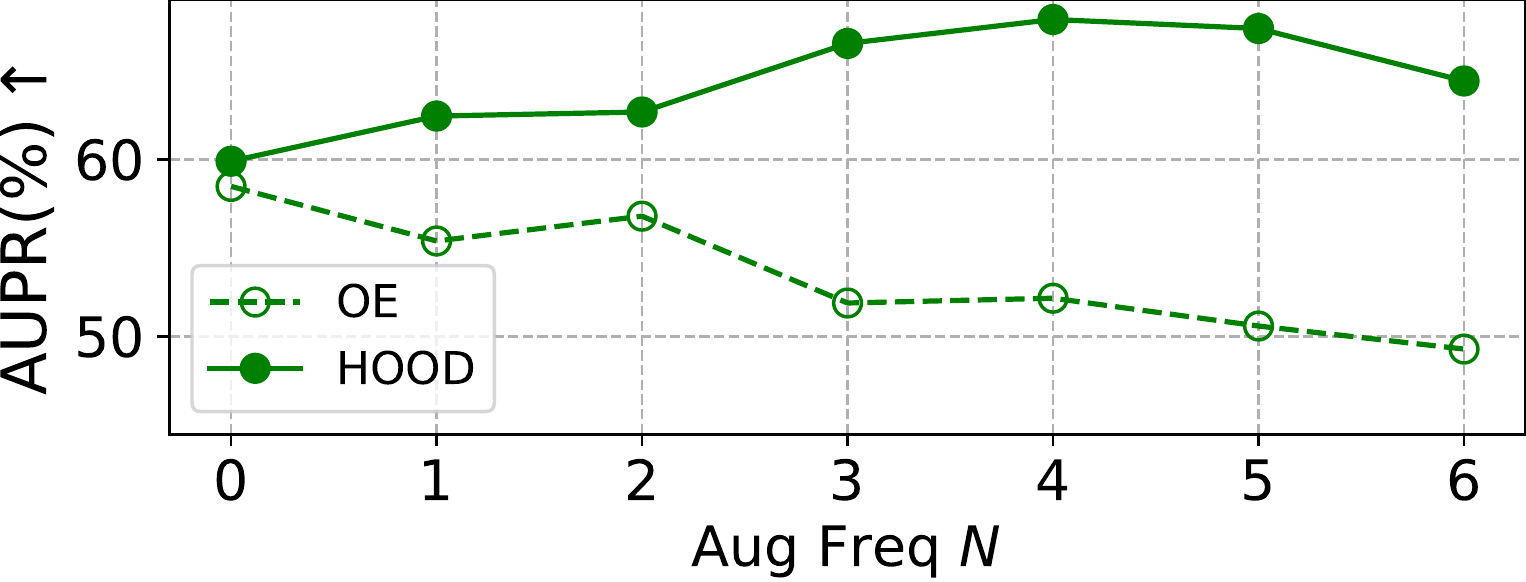}
        \caption{AUPR v.s. RandAugment frequency $N$}
        \label{fig:augn3}
    \end{subfigure}

    \caption{Ablation studies against RandAugment frequency $N$ for OOD training samples.}
    \label{fig:augn}
\end{figure*}

\subsection{Empirical analysis}
In this section, we conduct an empirical analysis to examine the effect of the contributing components and hyperparameters of HOOD.

\subsubsection{Analysis on the OOD training dataset}
Although recent works ~\cite{hendrycks2018deep,Liuenergy2020} have shown the effectiveness of exposure to real OOD data, it remains unclear how the training OOD data distribution influence the final OOD detection performance. Thus, we investigate the influence of training OOD data distribution and the sample size of OOD training data. In particular, we use the same training and test protocols as used for producing Table \ref{tab:cifar100}.

 {\bf Investigating the influence of training OOD data distribution.} To obtain multiple training OOD datasets with different distributions, we divide original training OOD data uniformly into 8 splits that can reflect the different difficulty levels of the OOD detection task. Specifically, we first train a vanilla MSP model on the in-distribution data without exposure to any OOD training data. We then rank all the training OOD samples in descending order according to their maximum softmax prediction score across all classes. We sequentially split all of these reordered samples into 8 groups with equal sizes.  Each group then reflects a different level of difficulty on OOD detection tasks observed by the MSP softmax predictor. The higher the maximum softmax value is, the more difficult the deep model can differentiate the OOD sample from the inliers. We then train each baseline model by respectively using the 1-8 groups of OOD training data corresponding to such decreasing level of difficulty.  As Fig. \ref{fig:difficulty1}, \ref{fig:difficulty2}, \ref{fig:difficulty3} illustrate, the performance is bell-shaped w.r.t. such difficulty levels. HOOD performs the best on the $5^{th}$ and $6^{th}$ difficulty level, whereas the performance drops either when the test sample is too ``simple'' or too ``difficult''. This might be due to the nature of HSIC: if test sample itself is already showing evident independence (easy samples) from the inliers, then HSIC cannot help further. However, if the OOD training samples are inherently correlated with inlier data in high-level semantics (difficult samples), HSIC also can fail.

 {\bf Investigating the influence of training OOD data size.} We firstly randomly subsample training OOD data from the original dataset (i.e., 80 Million Tiny Images). In detail, we sample 1K, 10K, 100K, 1M, 5M, and 10M OOD samples from the complete training OOD data, respectively.
Fig.~\ref{fig:size1}, \ref{fig:size2}, \ref{fig:size3} show that HOOD is relatively less sensitive to such size than distribution, with a relatively flat fluctuation range against the final detection performance. Intuitively, HOOD achieves better performance when exposed to more outliers. However, as training OOD size decreases, the performance does not drop too much until the training outlier number becomes lower than 10K. This shows that HOOD is robust against the training OOD data size, which does not necessarily require large-scale training outliers to work. Most importantly, we notice that HOOD can achieve comparable performance when it is even trained on 0.1\% training OOD data (i.e., 10K OOD data).

\subsubsection{Impact of strong augmentation.} \label{sec:aug}
As mentioned in ~\cite{hein2019relu,RenLiklihood2019,sinha2021negative,tack2020csi}, strong augmentation including shifting transformation~\cite{tack2020csi} or image permuting~\cite{hein2019relu,RenLiklihood2019} can be used to generate OOD samples. We are then intrigued to investigate the influence of strong augmentation on various OOD detection methods. In Section M.5.4, we have briefly discussed the impact of strong augmentation respectively on HOOD and OE, where HOOD consistently benefits from such expanded distribution whereas OE suffers. Here, we further examine the influence of different augmentation strengths on HOOD and OE. Specifically, we adopt RandAugment~\cite{cubuk2020randaugment} as our strong augmentation tool. RandAugment has two hyperparameters (i.e., $N$ and $M$), where $N$ indicates the number of augmentation transformations repeatedly applied on each input, and $M$ presents the magnitude for each transformation. We firstly range over different $N$ values while we fix $M=10$. A larger $N$ refers to a larger number of augmentations repeated on each input, and correspondingly implies a stronger augmentation and more diverse distribution of OOD samples. We use $\sigma=5.0$ for this ablation study on HOOD.

\begin{table}[t]
	\begin{center}
	\caption{\small Ablation study on different statistical test techniques.}
	\begin{adjustbox}{width=0.49\textwidth}\setlength\tabcolsep{6pt}
        \begin{tabular}{lllllll}
        \shline
        \multirow{2}{*}{\textbf{Method}}    & \multicolumn{2}{c}{\textbf{FPR95(\%)}$\downarrow$} & \multicolumn{2}{c}{\textbf{AUROC(\%)}$\uparrow$} & \multicolumn{2}{c}{\textbf{AUPR(\%)}$\uparrow$}   \\
                                                                   & SFM                                    & COR                         & SFM                          & COR                    & SFM                       & COR                    \\ \hline
        MSP   \cite{hendrycks17baseline} & 65.84                                   & 80.03                        & 74.76                         & 65.03                   & 33.25                     & 25.74                   \\
        OE\cite{hendrycks2018deep}            & 47.61                                   & 40.72                        & 86.82                         & 86.81                   & 58.48                     & 57.67                   \\
        HOOD                                                & \textbf{45.05}                       & \textbf{34.44}           & \textbf{87.91}             & \textbf{88.37}       & \textbf{63.65}         & \textbf{67.90}       \\ \shline
    \end{tabular}
    \label{tab:ablatetest}
        \end{adjustbox}
        \end{center}
\end{table}

\begin{table}[t]
	\begin{center}
	\caption{\small Impact of different kernels.}
	\begin{adjustbox}{width=0.49\textwidth}\setlength\tabcolsep{6pt}
        \begin{tabular}{lccc}
        \shline
        \textbf{Method}                                                         & \textbf{FPR95(\%)$\downarrow$} & \textbf{AUROC(\%)$\uparrow$} & \textbf{AUPR(\%)$\uparrow$} \\ \hline
         HOOD  (RBF)                                                             & {\bf 34.44}     & 88.37     & {\bf 67.90}                  \\
         HOOD  (Linear)                                                             & 34.72 & 88.15 & 67.54                 \\
         HOOD  (IMQ)                                                             & 35.13 & {\bf 88.46} & 67.33                 \\
        \shline
        \end{tabular}
        \label{tab:kernel}
        \end{adjustbox}
        \end{center}
\end{table}

As Fig. \ref{fig:augn1}, \ref{fig:augn2}, \ref{fig:augn3} illustrate, the optimal range of $N$ for HOOD is between 1 and 4. In this range of $N$, all HOOD+aug models outperform the HOOD baseline (i.e., $N=0$), while the detection accuracy keeps improving until $N$ grows to 4. When $N$ is greater than 4, the performance starts to drop, and a stronger augmentation beyond $N=4$ becomes even detrimental to the performance. This is because an extremely large augmentation strength would make the noise dominant, while the noise itself would potentially distort the semantics of the input features. In contrast, OE+aug models always perform worse than their plain form (i.e., $N=0$), and the higher $N$ results in even worse performances. This phenomenon indicates that HOOD encourages better learning capability and generalization ability than OE.

\subsubsection{Analysis on the statistical test}
{\bf Ablation on different statistical test.}
HOOD consists of two important components: the training loss (i.e., Eq.(\ref{eq:overall})) and the proposed test during inference. We demonstrate that both of these two components can independently contribute to HOOD's performance superiority over baseline models. Table \ref{tab:ablatetest} reports the OOD detection performance with the same training protocol as used for Table \ref{tab:cifar100} (HOOD+aug). Column SFM displays the test performance of each OOD detection approach by conventionally thresholding the softmax prediction as defined for OE and MSP. Column COR instead reports the detection result by using our proposed test (as described in Section \ref{sec:test}) on various methods. Table \ref{tab:ablatetest} well illustrates that even if we test HOOD model by using the conventional softmax thresholding method, i.e., we examine the HOOD training loss Eq.(\ref{eq:overall}) alone, HOOD still outperforms that of baseline OE and MSP across all metrics (under HOOD's optimal hyperparameters). When we associate the HOOD training with the exclusively defined new test (Column COR), HOOD further improves its own FPR95 and AUPR metrics by a significant margin, while this new test can also help improve OE model's FPR95 metric in many cases.

{{\bf Impact of different kernels.} We further experimented by using alternative linear/non-linear training kernels to implement HSIC metric during training. In particular, we respectively adopt linear and inverse multi-quadratic (IMQ) kernels. Table~\ref{tab:kernel} illustrates that these alternative kernels perform similarly to the RBF kernel. Note that using ``linear kernel'' to compute the HSIC is still an independence measurement that resorts to the calculation of Eq. (\ref{eq:tracehsic}), and does not indicate simply a linear correlation calculation. Hence, it demonstrates that our HOOD framework is relatively robust to the selection of kernels during training.}

\begin{figure}[t]
	\begin{minipage}[t]{0.5\linewidth}
    \includegraphics[width=1.0\textwidth]{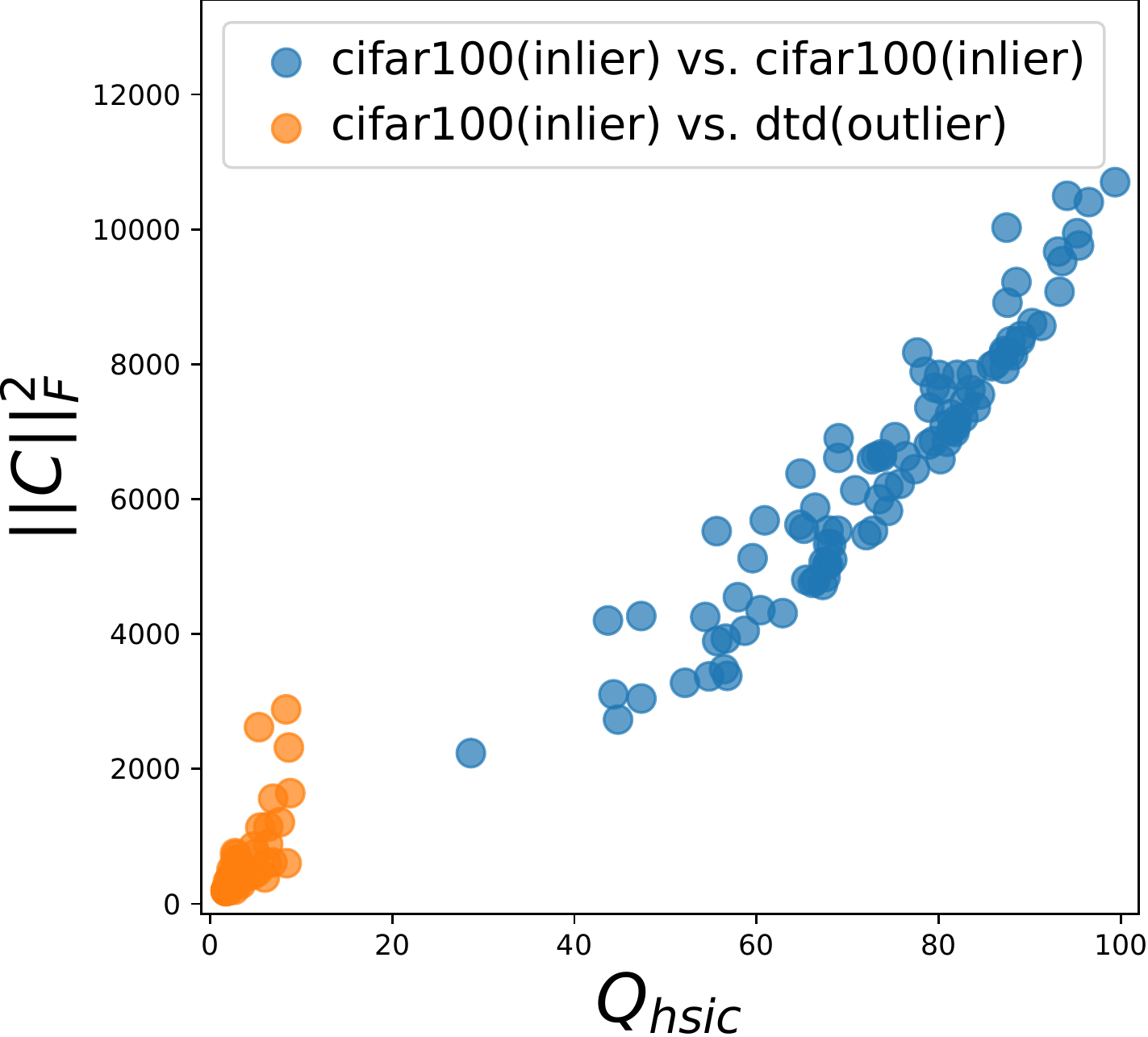}
    \caption{Empirical study on relation between $\mathcal{Q}_{hsic}$ and $\|\bb{C}\|^2_{F}$.}
    \label{fig:rel}
    \end{minipage}
    \hfill
    \begin{minipage}[t]{0.49\linewidth}
    \includegraphics[width=1.0\textwidth]{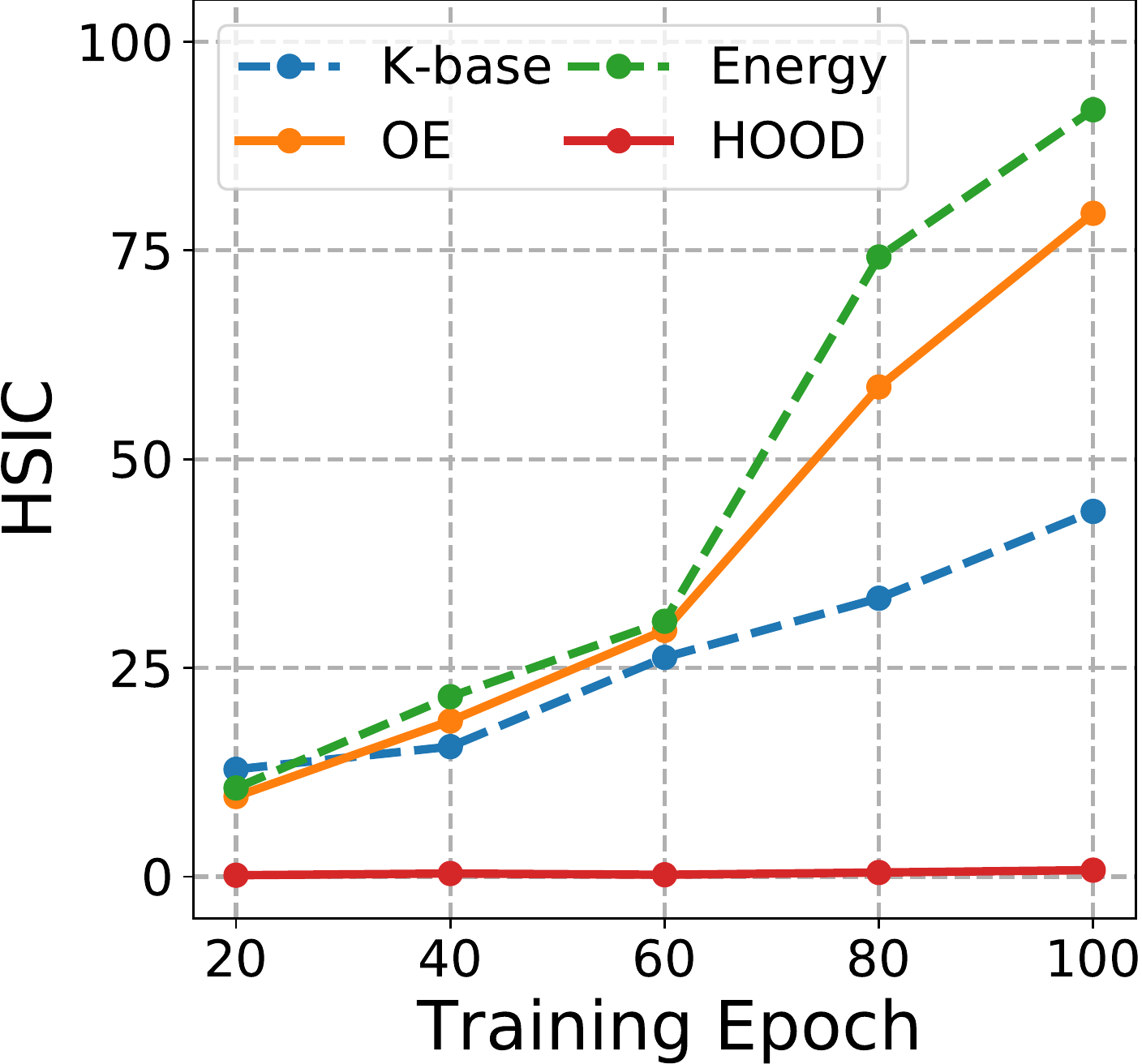}
    \caption{Empirical HSIC analysis during training.}
    \label{fig:hsic}
    \end{minipage}
\end{figure}

\begin{figure*}[t]
    \centering
    \begin{subfigure}{0.325\textwidth}
        \includegraphics[width=\textwidth]{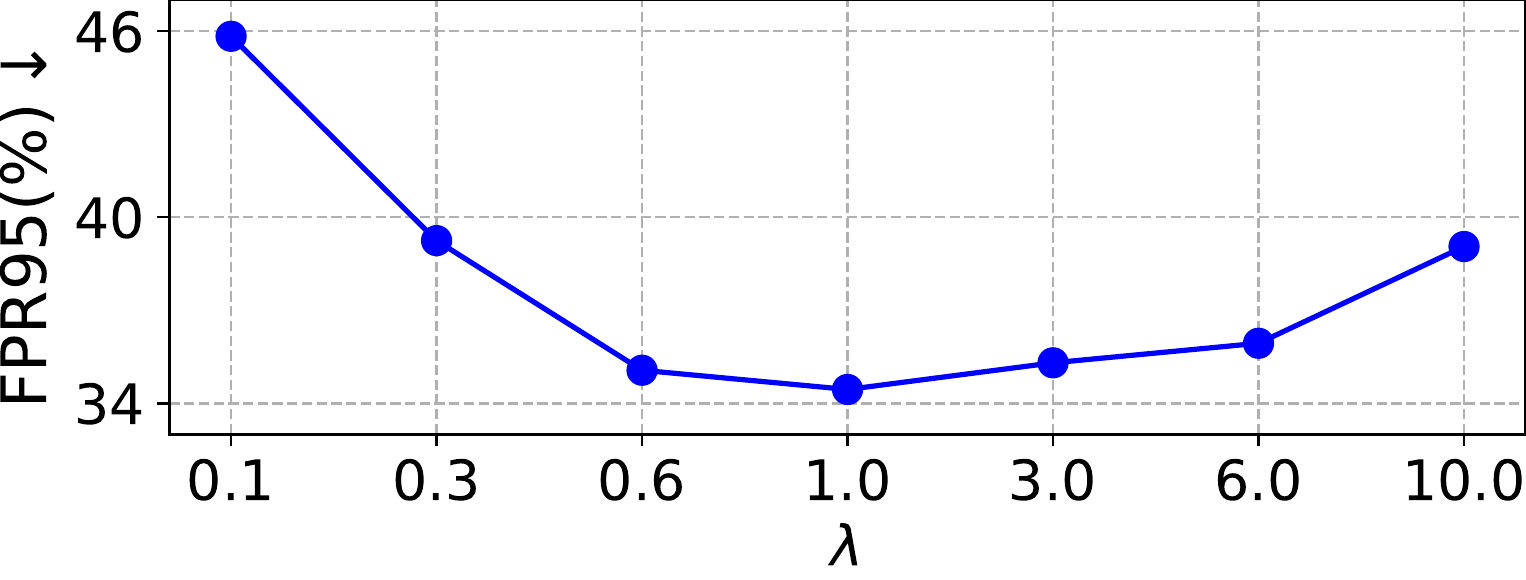}
        \caption{FPR95 v.s. HSIC loss weight $\lambda$}
        \label{fig:lambda1}
    \end{subfigure}
    \begin{subfigure}{0.325\textwidth}
        \includegraphics[width=\textwidth]{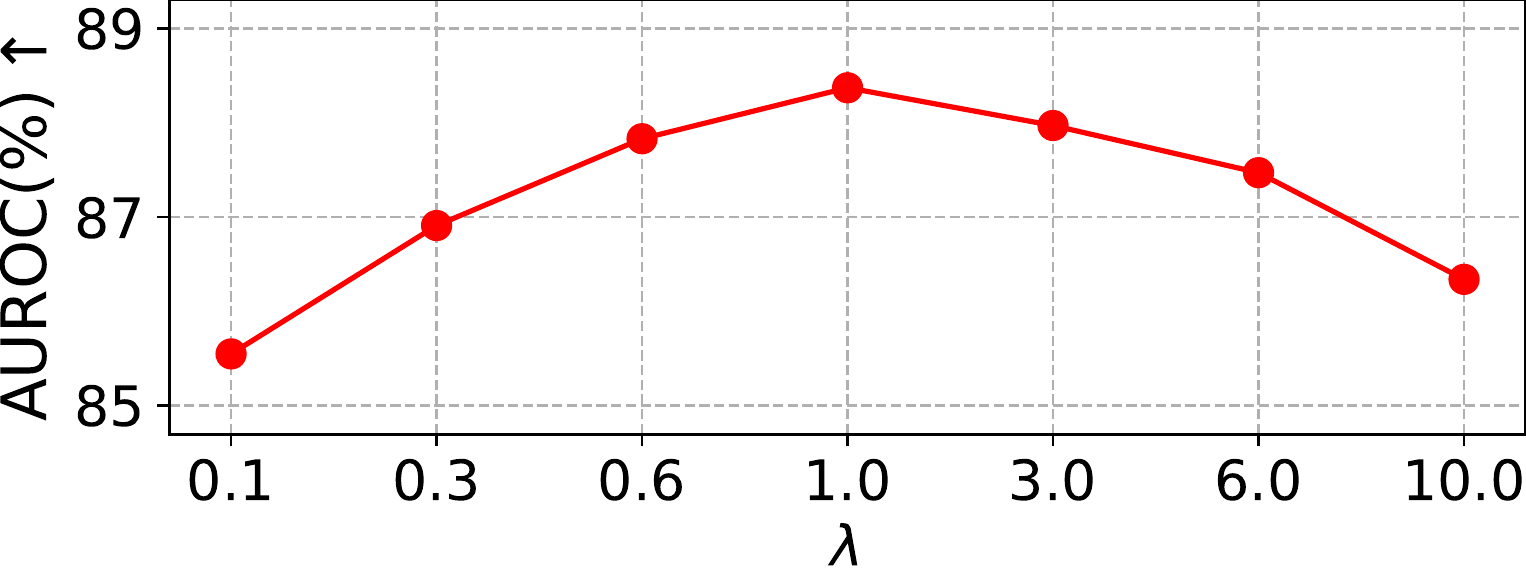}
        \caption{AUROC v.s. HSIC loss weight $\lambda$}
        \label{fig:lambda2}
    \end{subfigure}
    \begin{subfigure}{0.325\textwidth}
        \includegraphics[width=\textwidth]{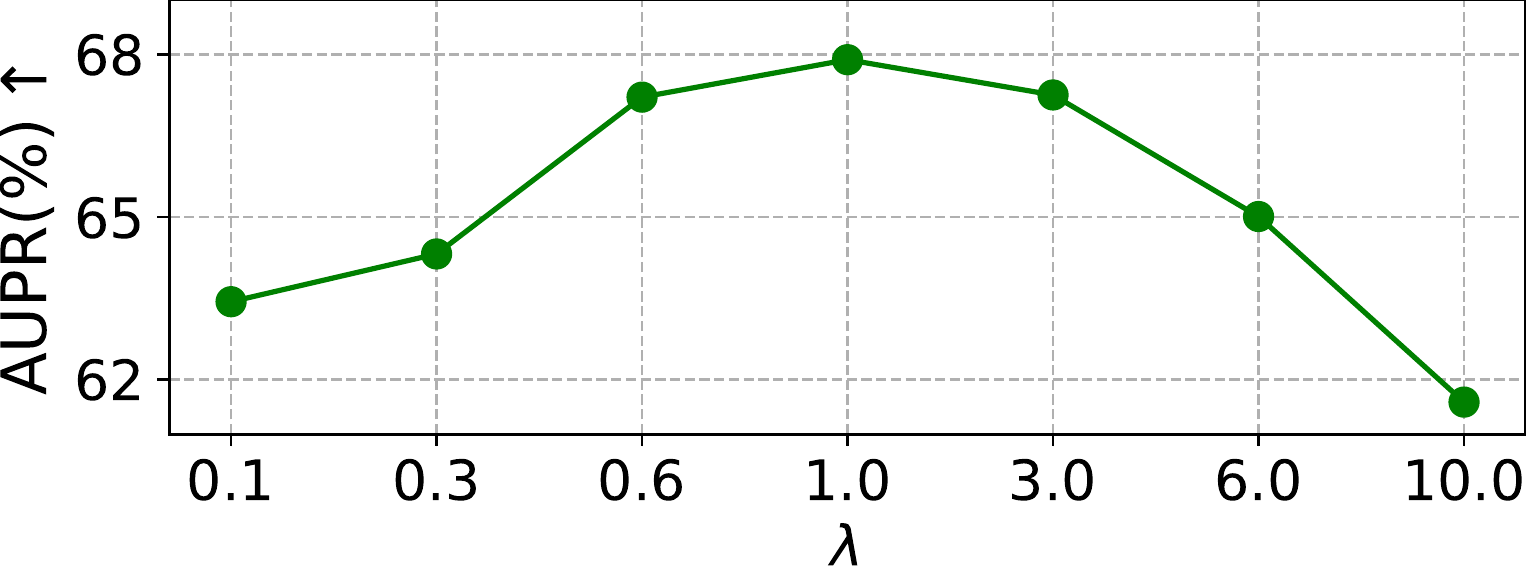}
        \caption{AUPR v.s. HSIC loss weight $\lambda$}
        \label{fig:lambda3}
    \end{subfigure}
    \\ \vspace{1em}
    \begin{subfigure}{0.325\textwidth}
        \includegraphics[width=\textwidth]{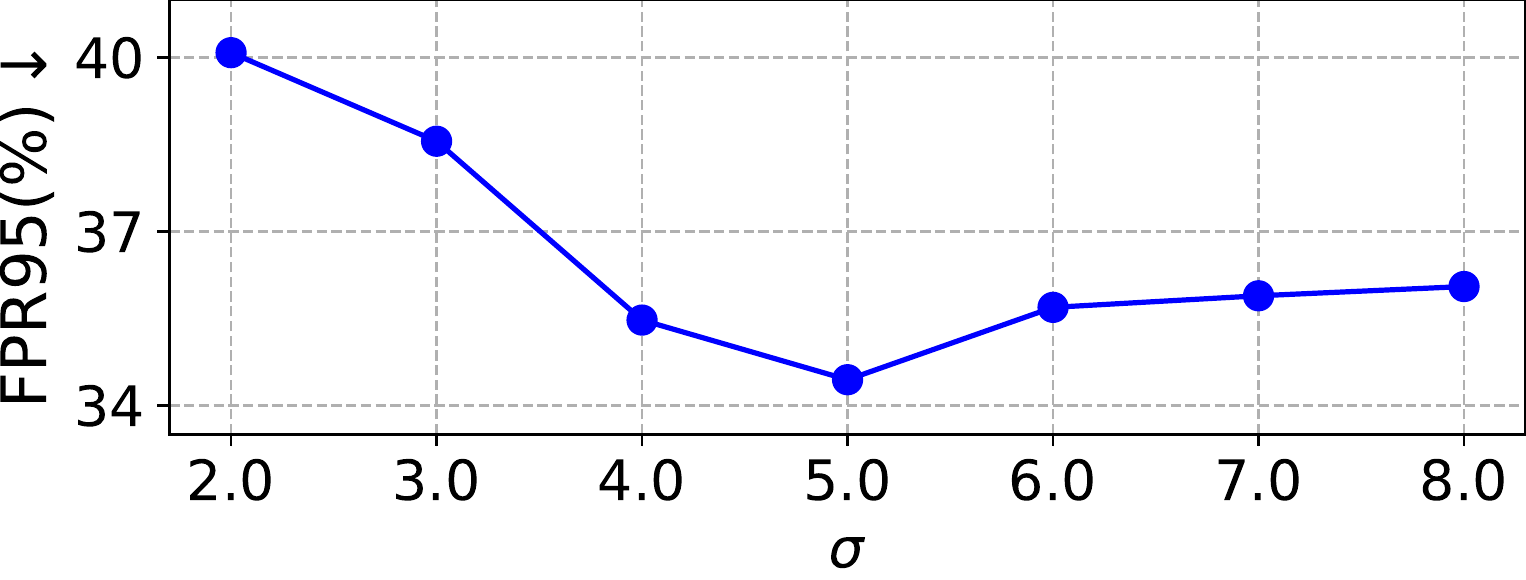}
        \caption{FPR95 v.s. RBF kernel temperature $\sigma$}
        \label{fig:sigma1}
    \end{subfigure}
    \begin{subfigure}{0.325\textwidth}
        \includegraphics[width=\textwidth]{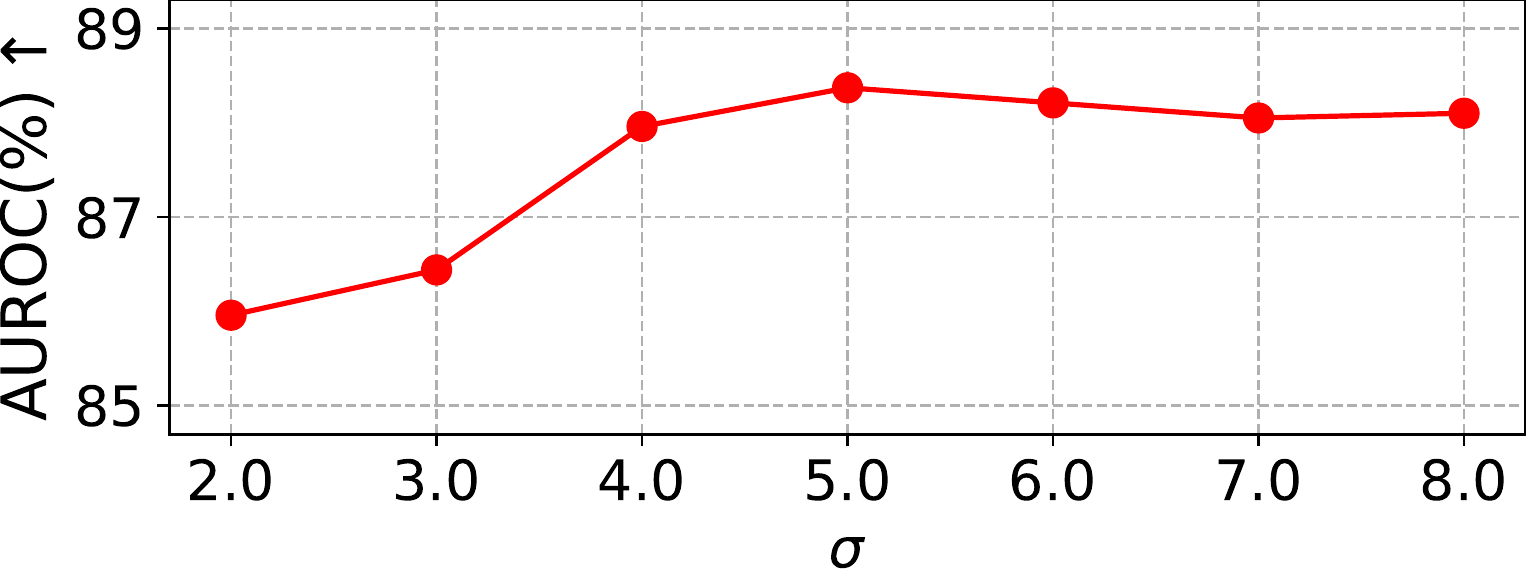}
        \caption{AUROC v.s. RBF kernel temperature $\sigma$}
        \label{fig:sigma2}
    \end{subfigure}
    \begin{subfigure}{0.325\textwidth}
        \includegraphics[width=\textwidth]{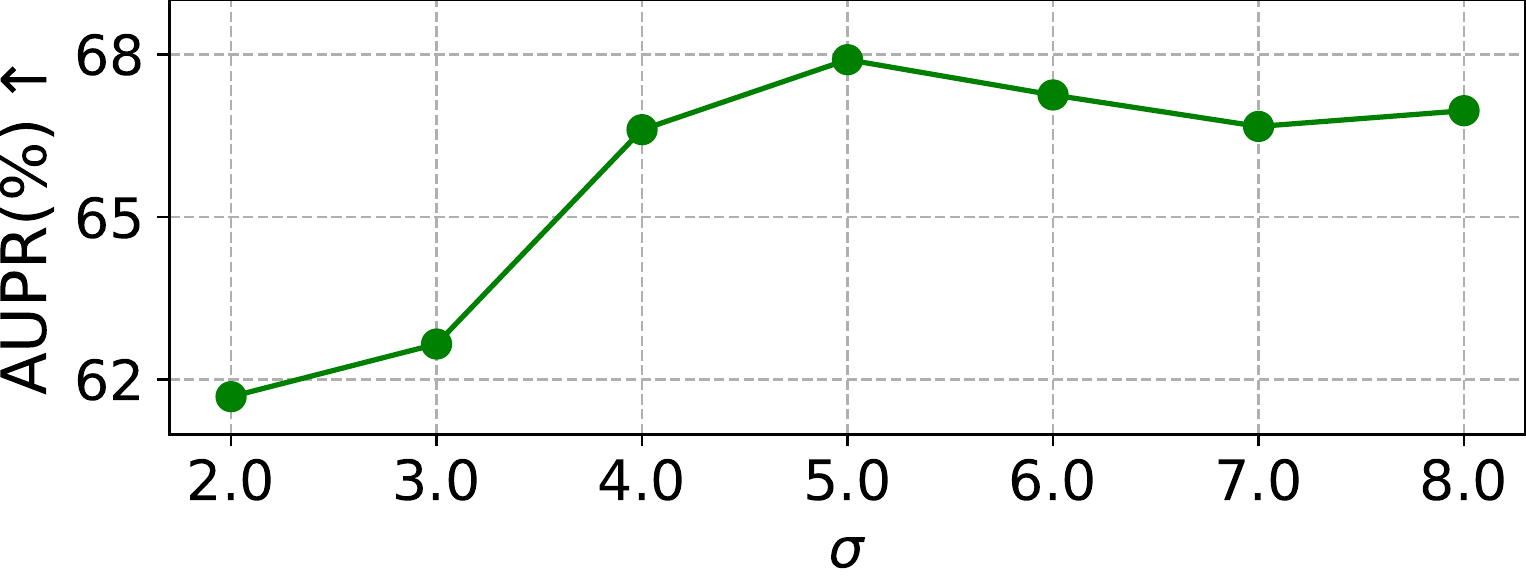}
        \caption{AUPR v.s. RBF kernel temperature $\sigma$}
        \label{fig:sigma3}
    \end{subfigure}
    \caption{Ablation studies against HSIC loss weight $\lambda$ and RBF kernel temperature $\sigma$.}
    \label{fig:abaltionall}
\end{figure*}

{\bf Correctness of our proposed statistical test.}
In Fig. \ref{fig:rel}, the x-axis value of each spot denotes the values of Eq. (\ref{eq:testhsic}) between a sample (either outlier or inlier) and a specific training class mean $\mu_c$. The y-axis measures the value in Eq. (\ref{eq:covariance}) where we repeatedly concatenate the same test sample $N$ times in constructing $\bb{G}$. We use the blue and orange colors to respectively represent the spots either using inlier (blue) and outlier (orange) in computing the $\bb{G}$ matrix. Fig. \ref{fig:rel} shows that our proposed test metric in Eq. (\ref{eq:testhsic}) indeed behaves like a lower bound of Eq. (\ref{eq:covariance}), and Eq. (\ref{eq:testhsic}) monotonically increases as the value of Eq. (\ref{eq:testhsic}) increases. It also shows good separation along the proposed test metric (x-axis) between inlier and outliers. Therefore, thresholding test metric under Eq. (\ref{eq:testhsic}) can correctly reflect the trend of HSIC metric, empirically validating our theoretical hypothesis and the efficacy of our proposed test.

\subsubsection{Mutual information metric comparisons during training}
The probabilistic paradigm through the HSIC aims to ensure that inlier data reveals little information about outliers. To support this claim, we provide quantitative measurements between the in-distribution and out-of-distribution that HOOD can indeed reduce the mutual information between the inliers and outliers. While it is broadly recognized that mutual information is hard to estimate, we instead illustrate in Fig.~\ref{fig:hsic} that HOOD can always incur the smallest HSIC loss between outliers and inliers when compared to other counterparts. This is based on the fact that mutual information is closely tied to the HSIC metric \cite{Grettonhsic2005}, i.e., two variables are independent (i.e., zero HSIC) if and only if their mutual information is zero.

\subsubsection{Ablation study on other hyperparameters}\label{sec:hyperparam}
We explore the impact of tuning the HSIC loss weight $\lambda$ and kernel temperature $\sigma$ in HOOD loss Eq.(\ref{eq:overall}).

\textbf{HSIC loss weight $\lambda$}. Fig. \ref{fig:lambda1}, \ref{fig:lambda2}, \ref{fig:lambda3} show that HOOD is able to maintain a performance superiority over a very broad range of $\lambda$ values, demonstrating the relative robustness of the HOOD method. Particularly, if $\lambda$ is drastically big, then the classifier on in-distribution data would be interrupted, as the HSIC loss would dominate the training, which is harmful to learning useful semantic features. However, if $\lambda$ becomes too small, then the impact of HSIC loss will diminish, leading to sub-optimal performance. We finally set HSIC loss weight $\lambda=1.0$.

\textbf{RBF kernel temperature $\sigma$}. In Fig. \ref{fig:sigma1}, \ref{fig:sigma2}, \ref{fig:sigma3}, we defer the the performance curve of HOOD against such variation of temperature $\sigma$, where $\lambda=1.0$. As can be seen, HOOD turns out to be relatively robust against the change of $\sigma$. In particular, the suitable range of $\sigma$ is reasonably wide for HOOD to work, and the optimal $\sigma$ is between 4 and 8, where we find the optimal $\sigma=5.0$ given the $\lambda=1.0$.

\section{Conclusion and Discussion}

In this paper, we propose a novel OOD detection paradigm named HOOD. We find the Hilbert-Schmidt Independence Criterion (HSIC) an intrinsically suitable optimization objective for the OOD detection task. HOOD capitalizes on the reduced mutual information between in-distribution data and OOD data by explicitly penalizing the HSIC metric between these two data groups. Empirical results show significant superiority over the existing SOTA methods, with the empirical improvement on many conventional evaluation metrics. We believe HOOD establishes a new and more efficient OOD detection paradigm without inducing extra computational complexity.


%

\ifCLASSOPTIONcaptionsoff
  \newpage
\fi



%

{\small
\bibliographystyle{IEEEtran}
\bibliography{egbib_pami}
}

\begin{IEEEbiography}[{\includegraphics[width=1in,height=1.4in,clip]{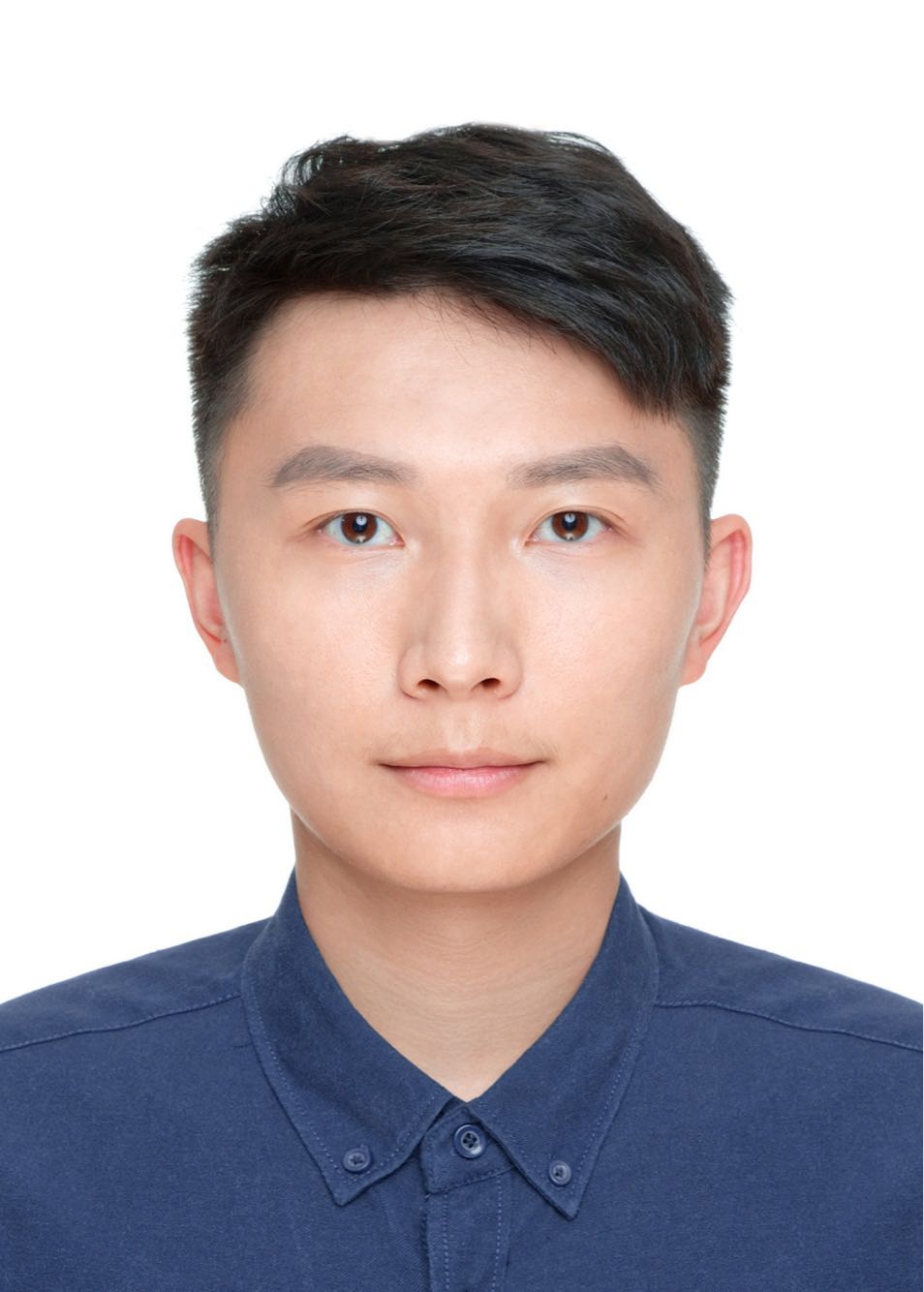}}]{Jingyang Lin}
 received the B.E. degree in 2019 from Sun Yat-sen University (SYSU), Guangzhou, China. He is currently a master student in the School of Computer Science and Engineering, SYSU. His research interests include computer vision, deep learning, self-supervised learning.
\end{IEEEbiography}

\begin{IEEEbiography}[{\includegraphics[width=1.2in,height=1.35in,clip,keepaspectratio]{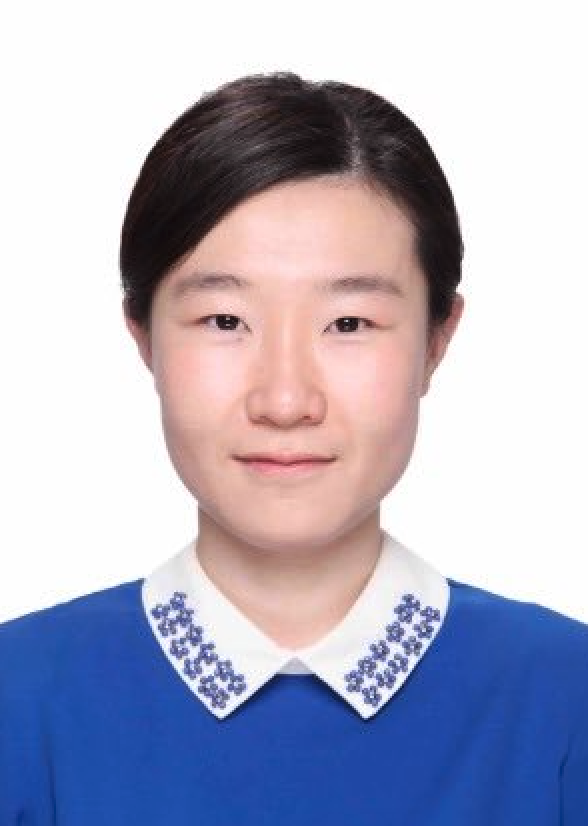}}]{Yu Wang} is currently a Senior Research Scientist at Qiyuan Lab, Beijing, China. Before joining Qiyuan Lab, she was a Researcher in Vision and Multimedia Lab at JD AI Research. Yu Wang achieved her PhD degree in Computer Science at the Department of Computer Science, University of Cambridge, UK in year 2016. She then joined the Department of Pure Mathematics and Mathematical Statistics, University of Cambridge, UK as a Postdoctoral Researcher during year 2016-2018. Her research interests include deep generative models, unsupervised learning, self-supervised learning, sparse coding, Bayesian inference, sparse Bayesian learning, and feature generalization.

\end{IEEEbiography}

\begin{IEEEbiography}[{\includegraphics[width=1in,height=1.3in,clip]{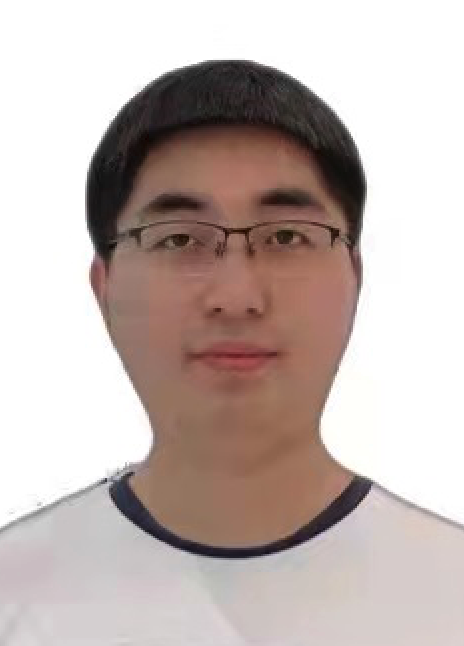}}]{Qi Cai} received the Ph.D. degree in 2022 from University of Science and Technology of China (USTC), Anhui, China. He developed a series of image understanding models with constrained annotations. He was also one of core designers of top-performing cross domain systems in worldwide competitions such as VisDA-2018 and VisDA-2019. His research interests include few-shot leaning, cross-domain detection and unsupervised pre-training.
\end{IEEEbiography}

\begin{IEEEbiography}[{\includegraphics[width=1in,height=1.25in,clip]{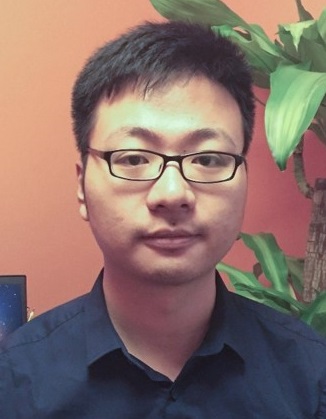}}]{Yingwei Pan}
is currently a Researcher in Vision and Multimedia Lab at JD AI Research, Beijing, China. His research interests include vision and language, domain adaptation, and large-scale visual search. He was one of core designers of top-performing multimedia analytic systems in worldwide competitions such as COCO Image Captioning, Visual Domain Adaptation Challenge 2018, ActivityNet Dense-Captioning Events in Videos Challenge 2017, and MSR-Bing Image Retrieval Challenge 2014 and 2013. He received Ph.D. degree in Electrical Engineering from University of Science and Technology of China in 2018.
\end{IEEEbiography}

\begin{IEEEbiography}[{\includegraphics[width=1in,height=1.25in,clip]{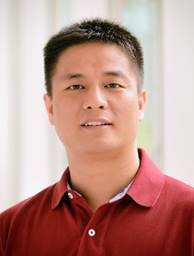}}]{Ting Yao}
is currently a Principal Researcher in Vision and Multimedia Lab at JD AI Research, Beijing, China. His research interests include video understanding, vision and language, and deep learning. Prior to joining JD.com, he was a Researcher with Microsoft Research Asia, Beijing, China. Ting is the principal designer of several top-performing multimedia analytic systems in international benchmark competitions such as ActivityNet Large Scale Activity Recognition Challenge 2019-2016, Visual Domain Adaptation Challenge 2019-2017, and COCO Image Captioning Challenge. He is the leader organizer of MSR Video to Language Challenge in ACM Multimedia 2017 \& 2016, and built MSR-VTT, a large-scale video to text dataset that is widely used worldwide. His works have led to many awards, including ACM SIGMM Outstanding Ph.D. Thesis Award 2015, ACM SIGMM Rising Star Award 2019, and IEEE TCMC Rising Star Award 2019. He is also an Associate Editor of IEEE Trans. on Multimedia.
\end{IEEEbiography}

\begin{IEEEbiography}[{\includegraphics[width=1in,height=1.4in,clip]{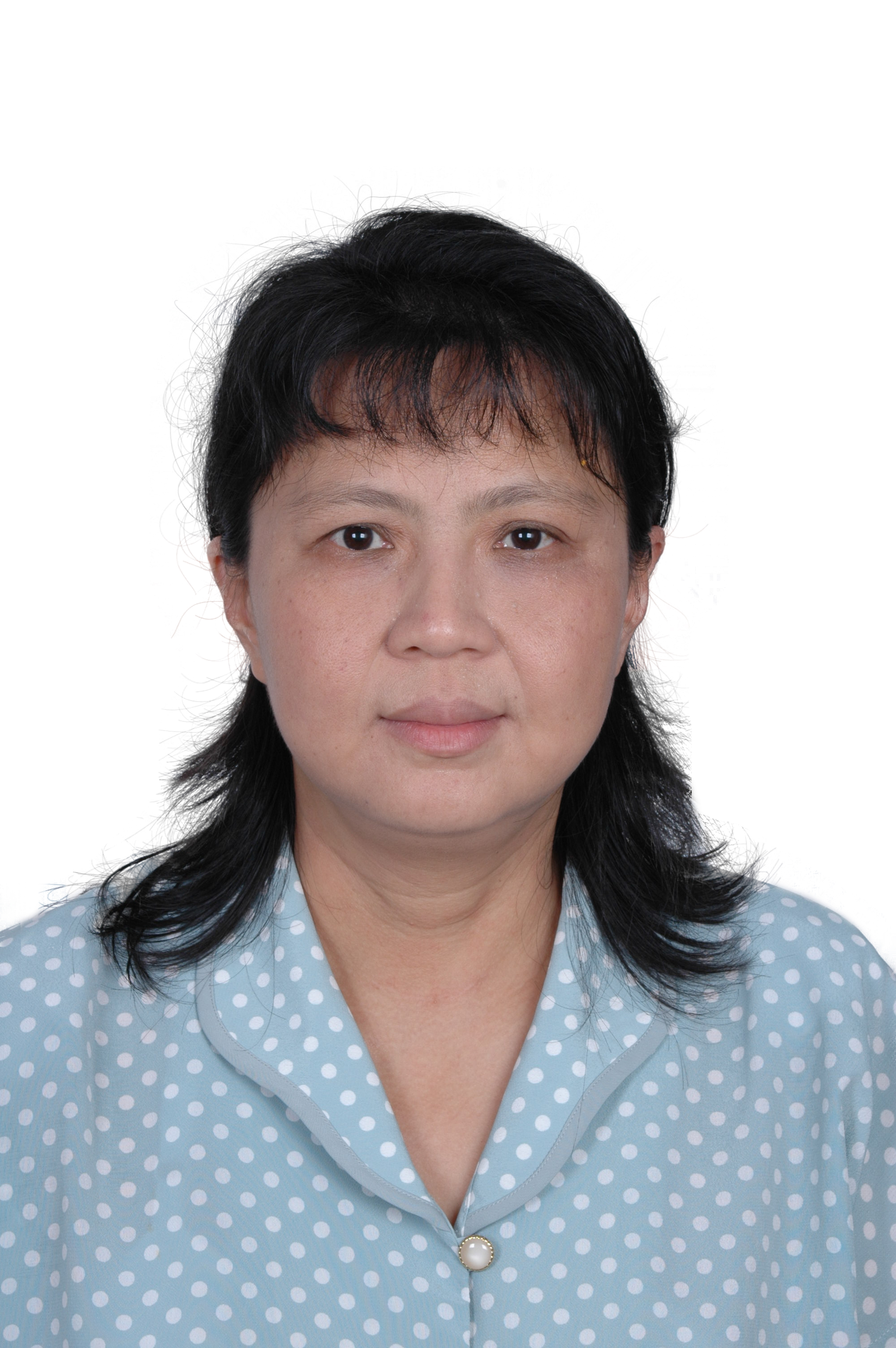}}]{Hongyang Chao}
received the B.S. and Ph.D. degrees in computational mathematics from Sun Yet-sen University, Guangzhou, China. In 1988, she joined the Department of Computer Science, Sun Yet-sen University, where she was initially an Assistant Professor and later became an Associate Professor. She is currently a Full Professor with the School of Data and Computer Science. She has published extensively in the area of image/video processing and holds 3 U.S. patents and four Chinese patents in the related area. Her current research interests include image and video processing, image and video compression, massive multimedia
data analysis, and content-based image (video) retrieval.
\end{IEEEbiography}

\begin{IEEEbiography}[{\includegraphics[width=1in,height=1.25in,clip,keepaspectratio]{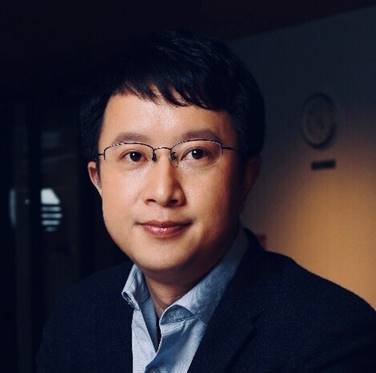}}]{Tao Mei}
(M07-SM11-F19) is a Vice President with JD.COM and the Deputy Managing Director of JD AI Research, where he also serves as the Director of Computer Vision and Multimedia Lab. Prior to joining JD.COM in 2018, he was a Senior Research Manager with Microsoft Research Asia in Beijing, China. He has authored or co-authored over 200 publications (with 12 best paper awards) in journals and conferences, 10 book chapters, and edited five books. He holds over 25 US and international patents. He is or has been an Editorial Board Member of IEEE Trans. on Image Processing, IEEE Trans. on Circuits and Systems for Video Technology, IEEE Trans. on Multimedia, ACM Trans. on Multimedia Computing, Communications, and Applications, Pattern Recognition, etc. He is the General Co-chair of IEEE ICME 2019, the Program Co-chair of ACM Multimedia 2018, IEEE ICME 2015 and IEEE MMSP 2015.

Tao received B.E. and Ph.D. degrees from the University of Science and Technology of China, Hefei, China, in 2001 and 2006, respectively. He is a Fellow of IEEE (2019), a Fellow of IAPR (2016), a Distinguished Scientist of ACM (2016), and a Distinguished Industry Speaker of IEEE Signal Processing Society (2017).
\end{IEEEbiography}

\clearpage

\appendices

\section{}

\subsection{Introduction}
The supplementary material contains:
{\bf 1)} We prove that penalizing HSIC in the form of linear kernel Eq.(\ref{eq:covariance}) can effectively penalize an upperbound of our Eq.(\ref{eq:testhsic});
{\bf 2)} The complete performance scores of Table \ref{tab:cifar100}, \ref{tab:places}, \ref{tab:tinyimage} regarding every specific OOD test dataset.

\subsection{Connection between Eq.(\ref{eq:covariance}) and Eq.(\ref{eq:testhsic})}
In this section, we prove that penalizing HSIC in the form of linear kernel Eq.(\ref{eq:covariance}) can effectively penalize an upperbound of our Eq.(\ref{eq:testhsic}) up to a scale. Define $\bb{C}$ to be cross-correlation matrix $\bb{C}=\bb{Z}^{\top}\bb{G}\in \mathbb{R}^{d\times d}$. Entries of $\bb{\mu}_c$ and $\bb{q}_{test}$ are denoted as $\mu_{i}$ and $q_i$. Notation $\bb{z}_{j} \in \mathbb{R}^{d\times 1}$ is the $j^{\text{th}}$ vector in $\bb{Z}^\top \in \mathbb{R}^{d\times N}$. We use $z_{i,j}$ to denote the $i^{\text{th}}$ entry in vector $\bb{z}_j$.

\begin{align}
|\bb{\mu}^{T}_c\bb{q}_{test}| &=\frac{1}{N}\bigg| \sum_j \bb{z}^{\top}_{j}\bb{q}_{test}\bigg| \\
&=\frac{1}{N}\bigg|    \sum_j \sum_i   z_{i,j}q_{i}  \bigg| \le \frac{1}{N} \sum_j \sum_i \bigg|z_{i,j}q_{i} \bigg|
\label{eq:upp}
\end{align}

where the inequality comes from the triangle inequality. In the meanwhile, if we concatenate the same test sample $\bb{q}_{test}$ multiple times to construct the columns of the matrix $\bb{G}$, it can be shown that Eq. (\ref{eq:covariance}) is equivalent to :
\begin{equation}
\|\bb{C}\|_F^2 = \sum_j \sum_i (z_{i,j}q_{i})^2 + \sum_j \sum_{m \neq n} (z_{m,j}q_{n})^2,
\label{eq:C}
\end{equation}
where the first term computes the on-diagonal correlation of $\bb{C}$ matrix, and the second term computes the off-diagonal cross correlation of $\bb{C}$ matrix, index $j$ ranges over multiple columns in $\bb{Z}^{\top}$ matrix. A zero-valued Eq. (\ref{eq:C}) forces both the first term $\sum_j \sum_i (z_{i,j}q_{i})^2$ and second term $\sum_j \sum_{m \neq n} (z_{m,j}q_{n})^2$ to be zero. Note penalizing the first term $\sum_j \sum_i (z_{i,j}q_{i})^2$ implies the upperbound $\sum_j \sum_i \bigg|z_{i,j}q_{i} \bigg|$  in Eq. (\ref{eq:upp}) is also zero. This connects our test Eq. (\ref{eq:testhsic}) with HSIC through the liner kernel construction Eq. (\ref{eq:covariance}). Note, although we constrain the $\bb{\mu}_c$ to be computed using each specific class, still if only a subset of in-distribution data shows strong dependence with test sample, then test is determined as an inlier. This does not interfere with our eventual motivation. The scale $\frac{1}{N}$ also does not interfere with the connection, since the constant scale does change the ranking of samples at test time.

\subsection{Full Performances}
For Table Table \ref{tab:cifar100}, \ref{tab:places}, \ref{tab:tinyimage} in the main paper submission, we have omitted the full performance results of each specific test dataset, where we have only reported the mean performance values averaged across different test datasets. Here in Table S.\ref{tab:full}, we report the complete performance scores of various baseline OOD detectors regarding each specific test datasets in Table \ref{tab:cifar100} (CIFAR-100), Table \ref{tab:places} (Place365) and Table \ref{tab:tinyimage} (TinyImageNet). For each $\mathcal{D}_{tr,in}$, we train each model with 5 different random seeds. Each trained model out of a single training run was tested over 10 different runs on OOD test sets. Given these results of multiple runs, we present mean and standard deviation ($\pm x$) over all OOD test datasets $\mathcal{D}_{tst,ood}$. As shown in Table \ref{tab:full}, the results demonstrate that HOOD successfully surpass OE and Energy methods in most majority cases, showing the good generalization ability of the HOOD method. It is also worthy to note that when the actual domain gap between $\mathcal{D}_{tr,in}$ and $\mathcal{D}_{tst,ood}$ is potentially small (e.g., $\mathcal{D}_{tr,in}=\text{CIFAR-100}$ vs. $\mathcal{D}_{tst,ood}=\text{CIFAR-10}$ or $\mathcal{D}_{tr,in}=\text{Place365}$ vs. $\mathcal{D}_{tst,ood}=\text{Place69}$), HOOD sometimes performs inferior to other methods. This encourages us to further extend the HOOD method against the fine-class OOD detection problem in our future work.

\begin{table*}[t] {
\caption{The full performance results of OOD detectors on each specific OOD test dataset for Table M.1 (CIFAR-100), M.2 (Place365), and M.3 (TinyImageNet). Test accuracy reported in \%.}
\begin{adjustbox}{width=1\textwidth}

\renewcommand\arraystretch{1.5}
\begin{tabular}{ll|cccc|cccc|cccc}
\shline
\multirow{2}{*}{$\mathcal{D}_{tr,in}$}           & \multirow{2}{*}{$\mathcal{D}_{tst,ood}$} & \multicolumn{4}{c|}{\textbf{FPR95(\%)$\downarrow$}}        & \multicolumn{4}{c|}{\textbf{AUROC(\%)$\uparrow$}}    & \multicolumn{4}{c}{\textbf{AUPR(\%)$\uparrow$}}      \\
                                               &                    & OE & Energy    & HOOD  & HOOD+aug & OE & Energy & HOOD & HOOD+aug & OE & Energy & HOOD & HOOD+aug \\ \hline

\multirow{5}{*}{\rotatebox[origin=c]{90}{\begin{tabular}[c]{@{}c@{}}\textbf{CIFAR-100}\\ (WideResNet)\end{tabular}}}  & DTD        & 48.14{\tiny$\pm$1.49}       & 51.46{\tiny$\pm$3.26}       & 53.97{\tiny$\pm$1.45} & {\bf 29.31{\tiny$\pm$2.75}} & 88.05{\tiny$\pm$0.36}       & 86.92{\tiny$\pm$0.85} & 88.56{\tiny$\pm$0.23}       & {\bf 94.64{\tiny$\pm$0.38}} & 64.41{\tiny$\pm$0.62}       & 62.67{\tiny$\pm$1.28} & 65.89{\tiny$\pm$1.50} & {\bf 82.51{\tiny$\pm$0.62}} \\
& SVHN       & 38.03{\tiny$\pm$5.74}       & 29.68{\tiny$\pm$6.84}       & 23.63{\tiny$\pm$3.98} & {\bf 7.02{\tiny$\pm$3.28}}  & 90.09{\tiny$\pm$1.76}       & 90.24{\tiny$\pm$3.33} & 94.06{\tiny$\pm$1.05}       & {\bf 98.36{\tiny$\pm$0.58}} & 62.78{\tiny$\pm$3.45}       & 60.30{\tiny$\pm$9.59} & 74.12{\tiny$\pm$4.71} & {\bf 89.72{\tiny$\pm$1.64}} \\
& Place365   & 46.45{\tiny$\pm$0.35}       & 37.90{\tiny$\pm$1.08}       & 34.41{\tiny$\pm$0.43} & {\bf 25.16{\tiny$\pm$0.59}} & 88.70{\tiny$\pm$0.15}       & 90.23{\tiny$\pm$0.25} & 91.06{\tiny$\pm$0.11}       & {\bf 94.40{\tiny$\pm$0.16}} & 64.69{\tiny$\pm$0.49}       & 65.82{\tiny$\pm$0.33} & 64.30{\tiny$\pm$1.14} & {\bf 78.09{\tiny$\pm$0.66}} \\
& LSUN       & 51.61{\tiny$\pm$0.82}       & {\bf 31.64{\tiny$\pm$1.10}} & 36.48{\tiny$\pm$2.11} & 31.97{\tiny$\pm$3.41}       & 87.19{\tiny$\pm$0.40}       & 91.44{\tiny$\pm$0.17} & 90.34{\tiny$\pm$0.44}       & {\bf 91.91{\tiny$\pm$0.98}} & 61.86{\tiny$\pm$0.70}       & 67.24{\tiny$\pm$0.47} & 60.63{\tiny$\pm$0.87} & {\bf 68.33{\tiny$\pm$2.50}} \\
& CIFAR-10   & {\bf 53.82{\tiny$\pm$0.19}} & 66.97{\tiny$\pm$1.16}       & 70.55{\tiny$\pm$2.46} & 78.73{\tiny$\pm$0.91}       & {\bf 80.06{\tiny$\pm$0.27}} & 71.29{\tiny$\pm$1.10} & 75.63{\tiny$\pm$0.38}       & 62.53{\tiny$\pm$1.12}       & {\bf 38.67{\tiny$\pm$0.75}} & 28.69{\tiny$\pm$1.20} & 34.57{\tiny$\pm$0.60} & 20.83{\tiny$\pm$0.45} \\ \hline
& {\bf Mean} & 47.61{\tiny$\pm$1.30}       & 43.53{\tiny$\pm$2.34}       & 43.81{\tiny$\pm$0.95} & {\bf 34.44{\tiny$\pm$1.12}} & 86.82{\tiny$\pm$0.43}       & 86.02{\tiny$\pm$1.05} & 87.93{\tiny$\pm$0.27}       & {\bf 88.37{\tiny$\pm$0.29}} & 58.48{\tiny$\pm$0.66}       & 56.94{\tiny$\pm$2.27} & 59.90{\tiny$\pm$0.71} & {\bf 67.90{\tiny$\pm$0.71}} \\ \shline

\multirow{4}{*}{\rotatebox[origin=c]{90}{\begin{tabular}[c]{@{}c@{}}\textbf{Place365}\\ (ResNet18)\end{tabular}}}    & DTD         & 79.94{\tiny$\pm$0.32} & 50.99{\tiny$\pm$2.22}       & 46.89{\tiny$\pm$4.08} & {\bf 30.99{\tiny$\pm$2.48}} & 74.35{\tiny$\pm$0.50} & 87.45{\tiny$\pm$0.80} & 83.22{\tiny$\pm$1.39}       & {\bf 92.14{\tiny$\pm$0.46}} & 32.80{\tiny$\pm$0.93}       & 56.37{\tiny$\pm$1.41} & 37.07{\tiny$\pm$2.20} & {\bf 65.05{\tiny$\pm$1.09}} \\
& SVHN        & 60.14{\tiny$\pm$6.47} & 9.23{\tiny$\pm$2.87}        & 10.96{\tiny$\pm$7.92} & {\bf 2.81{\tiny$\pm$0.60}}  & 85.61{\tiny$\pm$1.79} & 97.40{\tiny$\pm$0.87} & 97.03{\tiny$\pm$2.31}       & {\bf 99.30{\tiny$\pm$0.17}} & 56.85{\tiny$\pm$1.78}       & 83.59{\tiny$\pm$5.19} & 88.74{\tiny$\pm$7.88} & {\bf 95.00{\tiny$\pm$1.74}} \\
& Place69     & 89.15{\tiny$\pm$0.16} & {\bf 87.41{\tiny$\pm$0.39}} & 88.94{\tiny$\pm$1.30} & 91.22{\tiny$\pm$0.53}       & 62.30{\tiny$\pm$0.11} & 63.56{\tiny$\pm$0.32} & {\bf 64.35{\tiny$\pm$1.37}} & 58.80{\tiny$\pm$0.59}       & 24.19{\tiny$\pm$0.18} & 23.85{\tiny$\pm$0.12} & {\bf 24.34{\tiny$\pm$0.97}} & 20.21{\tiny$\pm$0.21} \\
& ImageNet-1K & 86.40{\tiny$\pm$0.17} & 85.44{\tiny$\pm$0.58}       & 68.62{\tiny$\pm$0.88} & {\bf 68.13{\tiny$\pm$0.50}} & 70.11{\tiny$\pm$0.26} & 72.74{\tiny$\pm$0.25} & 80.01{\tiny$\pm$0.72}       & {\bf 83.80{\tiny$\pm$0.15}} & 40.00{\tiny$\pm$0.46}       & 41.34{\tiny$\pm$0.41} & 44.11{\tiny$\pm$1.24} & {\bf 53.90{\tiny$\pm$0.53}} \\ \hline
& {\bf Mean}  & 78.91{\tiny$\pm$1.60} & 58.27{\tiny$\pm$1.19}       & 53.85{\tiny$\pm$2.26} & {\bf 48.29{\tiny$\pm$0.79}} & 73.09{\tiny$\pm$0.38} & 80.29{\tiny$\pm$0.36} & 81.15{\tiny$\pm$0.96}       & {\bf 83.51{\tiny$\pm$0.15}} & 38.46{\tiny$\pm$0.61}       & 51.29{\tiny$\pm$1.56} & 48.57{\tiny$\pm$2.17} & {\bf 58.54{\tiny$\pm$0.54}} \\ \shline

\multirow{5}{*}{\rotatebox[origin=c]{90}{\begin{tabular}[c]{@{}c@{}}\textbf{Tiny-ImageNet}\\ (WideResNet)\end{tabular}}}  & DTD          & 11.56{\tiny$\pm$2.42}      & {\bf 0.00{\tiny$\pm$0.00}} & 0.01{\tiny$\pm$0.01}       & {\bf 0.00{\tiny$\pm$0.00}} & 97.99{\tiny$\pm$0.21} & 99.69{\tiny$\pm$0.06}       & 99.92{\tiny$\pm$0.04}       & {\bf 99.94{\tiny$\pm$0.05}}       & 96.14{\tiny$\pm$0.28} & 99.47{\tiny$\pm$0.07} & 99.73{\tiny$\pm$0.25}       & {\bf 99.84{\tiny$\pm$0.09}} \\
& SVHN         & 27.29{\tiny$\pm$6.12}      & {\bf 0.00{\tiny$\pm$0.00}} & 0.01{\tiny$\pm$0.01}       & {\bf 0.00{\tiny$\pm$0.00}} & 95.19{\tiny$\pm$2.41} & {\bf 100.0{\tiny$\pm$0.00}} & 99.99{\tiny$\pm$0.01}       & 99.99{\tiny$\pm$0.01}       & 87.95{\tiny$\pm$6.25} & {\bf 100.0{\tiny$\pm$0.00}}      & 99.70{\tiny$\pm$0.28}       & {\bf 100.0{\tiny$\pm$0.00}} \\
& Place365     & 0.01{\tiny$\pm$0.01}       & {\bf 0.00{\tiny$\pm$0.00}} & 0.01{\tiny$\pm$0.01}       & {\bf 0.00{\tiny$\pm$0.00}} & 99.76{\tiny$\pm$0.05} & {\bf 100.0{\tiny$\pm$0.00}} & 99.99{\tiny$\pm$0.01}       & {\bf 100.0{\tiny$\pm$0.00}} & 99.47{\tiny$\pm$0.12} & {\bf 100.0{\tiny$\pm$0.00}} & 99.71{\tiny$\pm$0.28}       & {\bf 100.0{\tiny$\pm$0.00}} \\
& LSUN         & {\bf 0.00{\tiny$\pm$0.00}} & {\bf 0.00{\tiny$\pm$0.00}} & 0.01{\tiny$\pm$0.01}       & {\bf 0.00{\tiny$\pm$0.00}} & 99.95{\tiny$\pm$0.01} & {\bf 100.0{\tiny$\pm$0.00}} & 99.99{\tiny$\pm$0.01}       & {\bf 100.0{\tiny$\pm$0.00}} & 99.88{\tiny$\pm$0.02} & {\bf 100.0{\tiny$\pm$0.00}} & 99.70{\tiny$\pm$0.28}       & {\bf 100.0{\tiny$\pm$0.00}} \\
& ImageNet-800 & {\bf 0.00{\tiny$\pm$0.00}} & {\bf 0.00{\tiny$\pm$0.00}} & {\bf 0.00{\tiny$\pm$0.00}} & {\bf 0.00{\tiny$\pm$0.00}} & 99.83{\tiny$\pm$0.04} & {\bf 100.0{\tiny$\pm$0.00}} & {\bf 100.0{\tiny$\pm$0.00}} & {\bf 100.0{\tiny$\pm$0.00}} & 99.65{\tiny$\pm$0.09} & {\bf 100.0{\tiny$\pm$0.00}} & {\bf 100.0{\tiny$\pm$0.00}} & {\bf 100.0{\tiny$\pm$0.00}} \\ \hline
& {\bf Mean}   & 7.77{\tiny$\pm$2.20}       & {\bf 0.00{\tiny$\pm$0.00}} & 0.01{\tiny$\pm$0.01}       & {\bf 0.00{\tiny$\pm$0.00}} & 98.54{\tiny$\pm$0.49} & 99.94{\tiny$\pm$0.01}       & 99.98{\tiny$\pm$0.01}       & {\bf 99.99{\tiny$\pm$0.01}} & 96.62{\tiny$\pm$1.28} & 99.89{\tiny$\pm$0.01}       & 99.77{\tiny$\pm$0.22}       & {\bf 99.97{\tiny$\pm$0.03}} \\ \shline\end{tabular}
\end{adjustbox}
\label{tab:full}}
\end{table*}

\end{document}